\RequirePackage{fix-cm}

\documentclass[twocolumn]{svjour3} 

\emergencystretch=\maxdimen
\smartqed
\usepackage{graphicx}

\usepackage[T1]{fontenc}
\usepackage{graphicx}
\usepackage{amsmath}
\usepackage{amssymb}
\usepackage[vlined,boxed,ruled]{algorithm2e}
\usepackage{array}
\usepackage{url}
\usepackage{float}
\usepackage{xcolor}
\usepackage{hyperref}
\usepackage{gensymb}
\usepackage{textcomp}
\usepackage{pdflscape}
\usepackage{subfigure}
\usepackage{booktabs}
\usepackage{siunitx}
\usepackage{makecell}
\usepackage{tabularx}
\usepackage{ulem}
\usepackage{booktabs}
\usepackage[toc,page]{appendix}
\usepackage{multirow}
\usepackage[misc]{ifsym}

\newcommand{\ie}{\textit{i.e.}, }
\newcommand{\eg}{\textit{e.g.}, }
\newcommand{\methodname}{COVID-LSTM}
\newcommand{\repo}{\url{https://github.com/geohai/covid-lstm}}

\def\rev#1{\textcolor{black}{#1}}

\definecolor{citecol}{rgb}{0,0,0.5}
\hypersetup{urlcolor=citecol,linkcolor=black,citecolor=citecol,colorlinks=true}

\begin{document}

\title{A spatiotemporal machine learning approach to forecasting COVID-19 incidence at the county level in the United States}

\titlerunning{Spatiotemporal machine learning for forecasting COVID-19}

\author{Benjamin Lucas 
        \and Behzad Vahedi 
        \and Morteza Karimzadeh 
        }

\institute{\hspace*{5mm} Benjamin Lucas \\
           \hspace*{5.25mm} benjamin.lucas@colorado.edu \\
           \\
           \hspace*{5mm} Behzad Vahedi \\
           \hspace*{5.25mm} behzad@colorado.edu \\
           \\
           \hspace*{0.5mm} \Letter \hspace*{0.5mm} Morteza Karimzadeh \\
           \hspace*{5.25mm} karimzadeh@colorado.edu \\
           \\
           \hspace*{5mm} Department of Geography \\
           \hspace*{5mm} University of Colorado Boulder, CO, USA.
}

\date{Received: date / Accepted: date}

\maketitle

\begin{abstract}
With COVID-19 affecting every country globally and changing everyday life, the ability to forecast the spread of the disease is more important than any previous epidemic.
The conventional methods of disease-spread modeling, compartmental models, are based on the assumption of spatiotemporal homogeneity of the spread of the virus, which may cause forecasting to underperform, especially at high spatial resolutions.
In this paper we approach the forecasting task with an alternative technique---spatiotemporal machine learning.
We present \methodname, a data-driven model based on a Long Short-term Memory deep learning architecture for forecasting COVID-19 incidence at the county-level in the US.
We use the weekly number of new positive cases as temporal input, and hand-engineered spatial features from Facebook movement and connectedness datasets to capture the spread of the disease in time and space.

\methodname\ outperforms the COVID-19 Forecast Hub's Ensemble model (COVIDhub-ensemble) on our 17-week evaluation period, making it the first model to be more accurate than the COVIDhub-ensemble over one or more forecast periods.
Over the 4-week forecast horizon, our model is on average 50 cases per county more accurate than the COVIDhub-ensemble.
We highlight that the underutilization of data-driven forecasting of disease spread prior to COVID-19 is likely due to the lack of sufficient data available for previous diseases, in addition to the recent advances in machine learning methods for spatiotemporal forecasting.
We discuss the impediments to the wider uptake of data-driven forecasting, and whether it is likely that more deep learning-based models will be used in the future.


\end{abstract}

\section{Introduction}
Since the first known case of COVID-19 in December 2019, the disease has grown into a pandemic of unprecedented scale, significantly impacting modern life in the twenty-first century.
There have been over 216 million confirmed COVID-19 infections globally as of August 2021, with the United States being particularly hard hit, accounting for over one-third of all infections and 636,000 deaths~\cite{Dong2020}.
As a result, forecasting the spread of the disease within the United States has been a significant focus of the Centers for Disease Control and Prevention (CDC) and National Institutes of Health (NIH).

The nation-wide spread of COVID-19 has necessitated continual adaptations in planning and response decisions.
While substantial uncertainty exists surrounding the continuing spread of COVID-19, a robust forecast can be used to inform policy, targeted interventions, and mitigation strategies.
During the pandemic, forecasts have been used to allocate medical resources that are in short supply (\eg ventilators, personal protective equipment, gowns, sanitizer) to the areas with high COVID risk~\cite{Chu2021,Cohen2020,Galanis2021}.
Moreover, they have influenced the assignment of travel nurses, a group of qualified nurses who are not employed at specific locations, but rather, have multiple short-term appointments at hospitals located anywhere in the US based on demand~\cite{Astin2021}.
Forecasts have also been shown to inform response and mitigation strategies~\cite{Wallinga2010,Lipsitch2011} and help identify preferred locations for vaccine efficacy trials~\cite{Dean2020}.
The decision of many government authorities to `lockdown' the population has been based upon the combination of the current incidence rate and short-term forecasting of COVID-19 incidence~\cite{Koh2020,Melnick2020}.
The importance of accurate forecasting is further underlined by the fact that tens of teams from a variety of academic areas and industries have shifted their research focus to COVID-19 forecasting since March 2020~\cite{Cramer2021}.

In March 2020, the University of Massachusetts Amherst created the COVID-19 Forecast Hub~\cite{Ray2020,Cramer2021}, and since then, has been publishing weekly forecasts of COVID-19 incidence and mortalities at the scales of national-, state-, and county-level.
In this work, we focus on forecasting incidence at the highest spatial resolution for which validation data is widely available, i.e., county-level, as the most important scale for planning, resource allocation, and medical equipment distribution. Additionally, adopting preventative or mitigation strategies such as business restrictions or mask mandates are largely implemented locally, based on county-level incidence and prevalence.

The majority of the forecasting surrounding the spread of COVID-19 has been produced using compartmental models from epidemiology, in particular SEIR models~\cite{Zou2020,Friedman2021,He2020,Radulescu2020}.
These models divide the population into compartments---such as Susceptible to the virus (S), Exposed to the virus (E), Infected with the virus (I), or Recovered from the virus (R)---and then, use the characteristics of the virus and population to estimate the flow of proportions between these categories in order to forecast the spread and/or duration of an epidemic.
These models have produced reasonable forecasts for many decades and different epidemics~\cite{Viboud2018,Johansson2019,Reich2019}, but their main strength lies in providing a framework for characterizing the reproduction rate of a disease, \ie the expected number of secondary cases produced by a single (typical) infection in a completely susceptible population.
When it comes to forecasting, their performance is undermined by the underpinning assumption of the spatiotemporal homogeneity of the spread of the virus~\cite{Getz2019,Ansumali2020}. 

In this paper, we propose an alternate approach to incidence forecasting by implementing a data-driven framework based on a spatiotemporal deep learning architecture, which we call \methodname, to reflect our adoption of Long Short-Term Memory (LSTM) network architecture.
Our method utilizes human movement and county connectedness data, published by Facebook and acquired from mobile devices carrying the Facebook app, to derive features quantifying the spread of the virus between counties.
We integrate these hand-engineered spatiotemporal features into a LSTM deep learning model for multivariate time series.
Our results demonstrate that this method is the first to be more accurate on average than the COVID-19 Forecast Hub's Ensemble model (COVIDhub-ensemble) at predicting COVID-19 incidence over multiple forecast horizons.

The main contributions of this paper can be summarized as follows:
\begin{enumerate}
    \item \methodname: A novel framework for integrating spatial features and temporal incidence data using spatiotemporal deep learning for disease spread forecasting;
    \item The first model to produce more accurate forecasts, on average, than the COVIDhub-ensemble at multiple forecast horizons;
    \item A novel use of Facebook's Social Connectedness Index and Movement Range datasets to define the strength of spatial connections between counties in the United States, and the amount of inter-county and intra-county population movement; and,
    \item Openly providing code and data for reproducibility, wider implementation, and future research.
\end{enumerate}

The remainder of this paper is set out as follows: in Section~\ref{related_work}, we discuss the COVID-19 Forecast Hub and present both compartmental models and other existing data-driven approaches; in Section~\ref{data}, we discuss the case data and explain our use of Facebook's Social Connectedness Index and Movement Range datasets to derive spatial features; in Section~\ref{model}, we present \methodname\ in detail; in Section~\ref{exp}, we present our experiments and comparisons with the leading models currently used by the CDC and NIH, and discuss some of the model parameters; in Section~\ref{discussion}, we comment on our forecasts, the methods, and the future of COVID-19 forecasting; and finally, we draw conclusions in Section~\ref{conclusion}.

\section{COVID-19 forecasting and related work} \label{related_work}
Severe Acute Respiratory Syndrome Coronavirus 2 (SARS-CoV-2) is the highly-contagious virus causing the COVID-19 respiratory illness, and responsible for the COVID-19 pandemic.
It is a relative of SARS-CoV-1, the disease that was responsible for the SARS epidemic in 2003-2004, and which initially provided valuable insight to the potential spread of this disease~\cite{Wang2020}.

\rev{This section gives an overview of COVID-19 disease-spread modeling, including the role of both compartmental models and machine learning models.
We discuss some of the highest performing models used to forecast the spread of COVID-19, with an emphasis on models addressing the same problem we are---forecasting county-level incidence in the US.
We also discuss the COVID-19 Forecast Hub and the resulting COVIDhub-baseline and COVIDhub-ensemble models, as used by CDC, which we later use for comparison in our experiments.}

\rev{\subsection{COVID-19 modeling}
The international impact of COVID-19 has lead to significant research efforts being invested into modeling various aspects of the disease and policy responses.
In a comprehensive review of COVID-19 modeling~\cite{Cao2021}, the authors identify over 200k published articles on COVID-19, with approximately 22k of them specifically related to modeling.
COVID-19 modeling publications cover a large range of aspects of the pandemic including disease spread~\cite{Xu2020,Zou2020,Rodriguez2020,Bandyopadhyay2020,Pal2020,Gautam2021}, transmission dynamics~\cite{Hong2020,Wang2020inference,Zhou2020,Kontis2020}, diagnosis~\cite{Das2020,Mukherjee2021}, contact tracing~\cite{Keeling2020,Ibrahim2021}, medical treatment~\cite{Kontis2020,Beck2020}, non-pharmaceutical interventions~\cite{Brinati2020,Dehning2020,Flaxman2020,Fang2020}, and socioeconomic influence and impact~\cite{Pichler2020,delrio2020,Li2020,Walker2020}.}

\rev{The work dedicated to modeling the spread of the disease can be categorized into two strategies---compartmental models and data-driven models---which approach the problem from different directions.
Compartmental models are based on characteristics of the disease while data-driven models learn the pattern and rate of spread through previously observed data.}

\subsection{Compartmental models}
Compartmental models are based on a conventional mathematical modeling technique for predicting the spread of infectious diseases.
They stratify the population into compartments depending on their relationship with the disease in question.
The basis for each of these models is the SIR model, which was developed in the early twentieth century, and assigns members of the population into one of three categories: \textit{S}usceptible to the virus, \textit{I}nfected with the virus, or \textit{R}ecovered from the virus~\cite{Kendall1956}.
The flow of members of the population from one state to the next is modeled by the following set of differential equations:

\begin{enumerate}
    \item[] \hspace{3.5em} $\frac{dS}{dt} = - \frac{\beta I S}{N}$ \\
    \item[] \hspace{3.5em} $\frac{dI}{dt} = \frac{\beta I S}{N} - \gamma I$ \hspace{6.5em} $(1)$ \\
    \item[] \hspace{3.5em} $\frac{dR}{dt} = \gamma I$ \\
\end{enumerate}

where S is the number of subjects in the population susceptible, I is the number of subjects currently infected, R is the number of subjects recovered, and N is the total population size.
Parameters $\beta$ and $\gamma$ are based on characteristics of the disease in the population, and calculate the proportion of the susceptible population that are becoming infected with the disease and the rate at which people recover from the disease, respectively.
The ratio of $\beta$ to $\gamma$, referred to as the reproduction rate or $R_0$, is the expected number of new infections for each individual infection in a completely susceptible population.
An $R_0$ of greater than 1 represents a disease that is growing in the community, while an $R_0$ of less than 1 represents a disease that is declining in incidence.

The classical SIR model can be extended by including additional compartments such as exposed (resulting in an S\textit{E}IR model) and/or deceased (SEIR\textit{D}), or other compartments specific to the disease~\cite{Hethcote2000}.
For COVID-19 modeling, an exposed category is useful to represent the population who are in proximity to someone with the disease, but are not yet showing symptoms due to the incubation period of the virus~\cite{IHME2020,He2020}.
In~\cite{Xu2020}, the authors categorize the community into 6 classes: Susceptible, Exposed, Infected, Quarantined, Insusceptible, and Recovered, resulting in a SEIQPR model.
The categories of this model do not follow a direct linear sequence either, \ie an exposed individual may progress to being infected, or they may pass to being quarantined, allowing for differing regulations between locations.
In~\cite{Zou2020}, the authors propose a SuEIR model, including a category for unreported or unconfirmed COVID-19 cases, as the number of cases reported in the US is believed to have largely been under-reported throughout the pandemic.
Other methods have added free parameters to compartmental models to account for government policy such as social distancing~\cite{Mwalili2020} and travel restrictions~\cite{Chinazzi2020}.

\rev{In~\cite{Gibson2020}, the authors build upon the traditional SEIRD model to facilitate real-time COVID-19 forecasting.
Their model, named the Mechanistic Bayesian Model (UMass-MechBayes), uses a nonparametric model of the transmission rate ($\beta_t$) against time.
This allows for the transmission rate to increase or decrease for each measurement period.
A similar approach is presented in~\cite{Arik2021}, where researchers use machine learning models to accurately and dynamically quantify the transitions between model compartments.
Both this model, named the COVID-19 Public Forecast model (Google\_Harvard-CPF), and the UMass-MechBayes model have been identified as producing highly accurate county-level forecasts in the US~\cite{Cramer2021}, and therefore, we use them as comparison models in Section~\ref{exp}.
}

\subsection{Data-driven approaches}
As the COVID-19 epidemic developed into a pandemic during 2020, it also provided data at a scale unprecedented in epidemiology.
Within the United States, incidence and death data has been recorded at the county-level, and made freely available on a daily basis.
Data of this magnitude has provided an opportunity for re-examining the conventional forecasting methods, and devising data-driven forecasting.

The wealth of data collected during the COVID-19 pandemic (as a result of its importance as well as the disease prevalence) provides an opportunity for researchers to use autoregressive processes and machine learning as alternative approaches to compartmental models.
Machine learning is data-driven, meaning that the models identify, and learn from, underlying spatiotemporal trends in the data.
Alternately, compartmental modeling is based on the assumptions of the spatiotemporal homogeneity and the homogeneity of the population~\cite{Getz2019}---assumptions that may be incorrect in the case of COVID-19.
While models have been extended to include free parameters to account for demographic factors~\cite{Castillo1989,Leclerc2009}, dependence of transmission rates on time~\cite{Koelle2006}, and metapopulation structure~\cite{Lloyd2004,Watts2005}, this often ends up with a large number of parameters that must be calibrated for a given disease and population, which can introduce errors in incidence forecasting.

One potential reason that data-driven methods are not common--relative to compartmental models--in the history of disease forecasting, is the potential sparsity of the data.
For example, the World Health Organization states that around 8,000 people globally were infected with the SARS-CoV-1 outbreak in 2003, and only 8 of these were in the United States~\cite{Peiris2004}.
It is reasonable to assume that machine learning methods, which traditionally improve in performance proportionally to the quantity of data available~\cite{Halevy2009,Chen2017}, might have underperformed in forecasting this outbreak.
However, with millions of reported infections worldwide, data scarcity is not an issue for researchers using data-driven approaches to forecasting the COVID-19 pandemic.

With the global-scale disrupting effects of the COVID-19 pandemic, several research groups leveraging applied Artificial Intelligence have diverted their attention to COVID-19 forecasting. Additionally, recent research in machine learning for sparse data and transfer learning should further facilitate its adoption for disease forecasting, even for smaller epidemics~\cite{Shastri2021}.

\rev{In~\cite{Reinhart2021}, researchers present an autoregressive time series model (CMU-TimeSeries) that uses time series of both incidence and death data to make forecasts at high spatial resolutions, such as the county level.}

DeepCOVID~\cite{Rodriguez2020} is recognized as the first deep learning-based COVID-19 forecasting model published for US data.
It uses a multi-layer perceptron to produce state-level forecasts using cases, deaths and hospitalizations as a multivariate input.
As this was the first data-driven model published, it is notable that the resulting forecasts are highly comparable with the state-of-the-art compartmental models at the state-level~\cite{Cramer2021}.

Subsequent to DeepCOVID, various studies have treated COVID-19 forecasting as a time series forecasting problem, and addressed it with deep learning.
The most popular is LSTM networks, which is also the architecture that we also utilize in this work, as they are the state of the art in learning patterns in temporal data.
LSTM networks are implemented for national-level COVID-19 forecasting in~\cite{Zeroual2020,Chimmula2020,Bandyopadhyay2020,Pal2020,Gautam2021,Zou2020}.
It is likely due to data sparsity, irregularity, and the required pre-processing, that many researchers have not used LSTM networks at smaller scales~\cite{Rodriguez2020}.
\rev{One notable model, however, is the Neural Relational Autoregressive Model by Facebook AI Research (FAIR-NRAR)~\cite{Le2021}.
This model uses temporal series of incidence, deaths, and other covariates quantifying the mobility of the population as input to a recurrent neural network.
This model is the most similar in structure to ours and is therefore included as a comparison model in our experiments in Section~\ref{exp}.}

\subsection{COVID-19 forecast hub} \label{subsec:forecasthub}
The COVID-19 Forecast Hub~\cite{Ray2020,Cramer2021}, created by the University of Massachusetts Amherst, is a repository of COVID-19 forecasts from various research groups across the US.
Each week, groups submit forecasts for the numbers of new cases, hospitalizations and deaths in future days, weeks, and months at the national, state, and county level in the United States.
The repository hosts over 100~million rows of data that are openly accessible.

The Forecast Hub works closely with the CDC and passes on forecasts for use in government communications.
The two common forecast models published by the CDC are the COVIDhub-baseline model and the COVIDhub-ensemble model; these are also the two models that we use for comparison in our experiments in Section~\ref{exp}.
The COVIDhub-baseline model's forecast is a neutral reference model with a predictive median equal to that observed over the same time period immediately prior.
In our experiments, this means that the predicted number of COVID-19 incidence for a given county during any future week will be equal to the number of reported infections in that county during the current week, \ie persistence.
The COVIDhub-ensemble model's forecast is a collaboration between the CDC, 21 research groups, five private industry groups, and two other government groups.
The forecast value is the median prediction out of all eligible models that are submitted through the COVID-19 Forecast Hub for a given forecast date, hence the size of the ensemble varies by week and location.
The number of individual models in the ensemble ranges from single figures during April 2020 to 49 during December 2020, however the forecasts for some locations include fewer models as not all model submissions contain predictions for all locations.
The COVIDhub-ensemble's forecast is rarely the most accurate for an individual point prediction, but is shown to be significantly better on average than any single model across all forecast horizons~\cite{Cramer2021}.

Forecast Hub predictions are published for various time horizons.
The focus of this work, and the majority of the model submissions, is on the more reliable `short-term' forecasts, which include forecasts for one, two, three, and four weeks into the future.
One important note is that the forecasts are made for epi-weeks, meaning that the forecast value for the two-week horizon is equal to the numbers anticipated to occur during the week (7 days) that is two weeks in the future, rather than a cumulative forecast for the coming two weeks (14 days).
Epi-weeks run Sunday through Saturday, as defined by the CDC, and are common practice in epidemiology.

Both a point prediction and a set of quantiles are necessary for each Forecast Hub submission to enable the creation of prediction intervals.
For forecasting incidence at the county-level, the published quantiles are 0.025, 0.1, 0.25, 0.5, 0.75, 0.9, and 0.975, and therefore, we have included these quantiles in our results (Section~\ref{quantiles}).

\section{Data} \label{data}
The task we aim to address is to predict COVID-19 incidence at the county-level in the contiguous US over 1-, 2-, 3-, and 4-week forecast horizons.
Our data has 10 temporal input features per instance: 2 derived from the raw number of cases; 6 features derived from Facebook datasets representing human movement and inter-county connectedness; and 2 weather features.
Each feature $x$ is included with a temporal lag of $n$ weeks: $[x_{t}, x_{t-1}, x_{t-2}, ..., x_{t-n}]$.

The input features are shown in Table~\ref{tab:features} and discussed further in the following subsections.

\begin{table*}[!htbp]
    \centering
    \caption{The features used in \methodname.}
    \label{tab:features}
    \begin{tabular}{c|l}
        \toprule
        \textbf{Source} & \textbf{Feature} \\
        \midrule
        \multirow{2}{*}{Johns Hopkins University CSSE~\cite{Dong2020}} & New Weekly COVID-19 Incidence \\
        & Monthly Mean cumulative COVID-19 Incidence \\
        \midrule
        \multirow{6}{*}{Facebook~\cite{herdagdelen2020,Bailey2018}} & Stay Put index \\
        & Rate of weekly change of Stay Put index \\
        & Change in Movement index \\
        & Rate of weekly change in Change in Movement index \\
        & Weekly change in Social Proximity to Cases \\
        & Monthly Mean Social Proximity to Cases \\
        \midrule
        \multirow{2}{*}{Weather} & Average Minimum Temperature \\
        & Average Maximum Temperature \\
        \bottomrule
    \end{tabular}
\end{table*}

\subsection{Reported cases}
\begin{figure}[htb!]
    \centering
    \includegraphics[width=.85\linewidth]{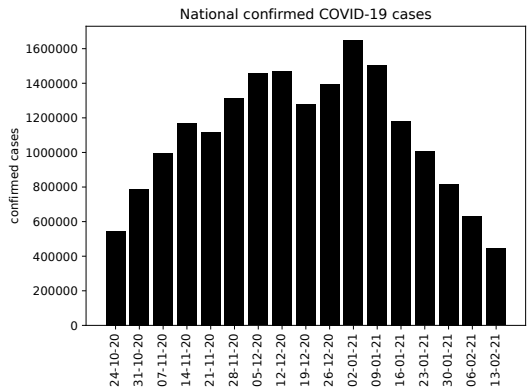}
    \caption{\label{fig:evaluation_period} The observed COVID-19 incidence for our 17-week evaluation period at the national-level. This period covers three periods of differential growth of the virus, a rapid increase in November and Early December, steady growth across the new year period, and a decline in growth from mid-January to February.}
\end{figure}

The raw data are downloaded from Johns Hopkins University's Center for Systems Science and Engineering~\cite{Dong2020} as the cumulative number of confirmed COVID-19 infections per county per day, \ie incidence.
We download data starting 1 April 2020 through 20 February 2021.
The period used for our evaluation begins on Saturday 31 October 2020 through 20 February 2021.
This evaluation period was chosen as it covers three different phases of the pandemic---the sharp increase in late 2020, the peak around the December-January holiday period, and the general decline beginning in the new year.
This is reflected in the national case numbers shown in Figure~\ref{fig:evaluation_period}.
All data between the start date and the given evaluation (\ie forecasting) date is used for training the forecast model, and all evaluation is performed on unseen (held out) data.

To process this data into a weekly dataframe, we first take a 7-day rolling average to smooth the irregularities caused by inconsistent reporting, especially on weekends and holidays.
From here, we calculate 2 input variables aligned with the epi-week-based forecast dates: (1)~the mean number of cumulative cases during a given week, and (2)~the increase in new cases reported during a given week \ie incidence.
This second variable is also the target variable for our model.

Although we have applied smoothing to the raw data, there are still a number of irregularities present.
To combat this, if a county reports a weekly increase of 0 cases, any input instance containing that value is excluded from the training data.
This assists the model to learn underlying temporal trends without the noise of inconsistent reporting.

We note that while our model is trained on a target of rolling average values, the final evaluation is compared to the raw number of reported cases in a given week (we discuss this further in Section~\ref{outliers}).

Our data contains the counties of 48 states in the contiguous US.
However, as we are using a time lag of length $l$, each county contributes multiple instances to the training set.
That is, for a single county's case data: $x_1, x_2, x_3, ..., x_t$, where we are trying to forecast $x_{t+1}$, the training data will include as individual instances:\\
\\
$x_1, x_2,..., x_{1+l} \rightarrow x_{2+l}$ \\
$x_2, x_3,..., x_{2+l} \rightarrow x_{3+l}$ \\
$x_3, x_4,..., x_{3+l} \rightarrow x_{4+l}$ \\
...\\
$x_{t-1-l}, x_{t-l},..., x_{t-1} \rightarrow x_{t}$ \\
\\
where the values to the right of the arrows are the target of the respective training instances. The test instance will be:\\
\\
$x_{t-l}, x_{t-l+1},..., x_{t} \rightarrow x_{t+1}$ \\

This means that the model is essentially learning patterns in multivariate temporal series of length $l+1$, for each county and time step.

The results presented in Section~\ref{exp} use input data with a temporal lag of 9.
We investigate the use of other temporal lags in Section~\ref{lag}.

\subsection{Facebook-derived spatial features}
In order to account for \textit{intra-county} human movement patterns, we included movement variables derived from the Facebook Movement Range dataset~\cite{herdagdelen2020}.
This anonymized, privacy-protecting dataset is generated by Facebook, and derived from mobile devices carrying the Facebook app, \ie by tracking the location of users' log-ins over time to measure the flow of the population. Within this dataset, there are two metrics, called (1) Change in Movement and (2) Stay Put.
Change in Movement is a measure of the relative change in aggregated movement within a county compared to a baseline of the month of February 2020, which is the month prior to the first cases of COVID-19 being recorded in the US.
Stay Put is a measure of the proportion of a county's population that have stayed within a small radius for a 24 hour period.
The Movement Range dataset is published daily, and we have calculated the change in movement and the rate of this change for the appropriate epi-week.

In order to account for \textit{inter-county} spread of the disease, we have also incorporated an index called Social Proximity to Cases (SPC)~\cite{Kuchler2021}, which is a COVID-19-specific metric incorporating Facebook's Social Connectedness Index (SCI)~\cite{Bailey2018}.
The SCI is a dataset published by Facebook that uses the general home location of Facebook friends to quantify the connectedness between those two administrative units (in our case, the locations are US counties, however SCI is not available at this scale in all locations internationally).
This means that two counties can be connected to one another without being spatially adjacent.
The Social Connectedness (SC) between two counties is calculated as the ratio of Facebook-friendships between users in those counties to the total number of possible Facebook-friendships between those counties, \ie SC represents the probability that any two Facebook users in different US counties are friends on Facebook, given their respective locations.
The SC between counties $i$ and $j$ would be:
\begin{equation} \tag{2}
    SC_{i,j} = \frac{FB\_friendships_{i,j}}{FB\_users_{i} \times FB\_users_{j}}
\end{equation}
The published value for SCI equals the value for SC scaled to a range of between 1 and 1,000,000,000 and rounded to the nearest integer.

In~\cite{Kuchler2021}, the authors found that SCI was highly correlated with the early spread of COVID-19 cases at the county-level in the US.
Their work defined the SPC metric to quantify the likely exposure of the population of a given county to positive cases from connected counties.
It is calculated as the weighted sum of the positive cases in the connected counties, where the weights are are the social connectedness between counties.
Specifically, the SPC for county $i$ at time $t$ is:

\begin{equation} \tag{3}
    SPC_{i,t} = \sum_{j \in C} Incidence\_rate_{j,t} \times \frac{SC_{i,j}}{\sum_{h \in C} SC_{i,h}}
\end{equation}

where $Incidence\_rate_{j,t}$ is the number of positive COVID-19 cases per 10,000 people in county $j$ at time $t$, and C is the set of all counties socially connected to county $i$ (in our case, C is the set of all other counties in the US).
SCI is a static index produced annually; however, as SPC is weighted by the weekly COVID-19 incidence rate, SPC is a dynamic measure of proximity to cases.

Together, these features help capture the heterogeneous spatial spread of COVID-19, both intra- and inter-county, for a given week~\cite{Vahedi2021}.
In order for our model to learn the spatiotemporal spread, we create temporal series of the weekly averages of each of these variables.
We also include a series of the rate of change of the variables (as the slope of a linear regression model) over that week, to account for average values that have high or low variance.

\subsection{Weather features}
There is evidence to suggest a correlation between climate and the spread of COVID-19~\cite{Malki2020,Tosepu2020,Bashir2020}, and as such, we have included temperature as a feature in \methodname.
The features included are the weekly average minimum temperature and weekly average maximum temperature, per county.
The averaging is performed only for the populated areas of each county based on the US Census's Populated Places.
As above, these variables are used as a temporal sequence of $l$+1 weeks in length.

\section{\methodname} \label{model}
Our \methodname\ model is an ensemble regression model based on a Stacked Long Short-Term Memory (LSTM) deep learning architecture~\cite{Hochreiter1997}, which is a type of recurrent neural network (RNN).
LSTM networks incorporate multiple loop connections, which enable information to be retained and flow from one point of the network to the next.
As a result, they have found much success in sequential and list-based data, such as speech recognition, image captioning, and time series forecasting, as in our application~\cite{Graves2008,Gers2000,Sak2014}.
In this section we will outline the \methodname\ architecture in depth, and discuss its hyperparameters.

\subsection{Architecture}
\begin{figure}[htb!]
    \centering
    \includegraphics[width=.95\linewidth]{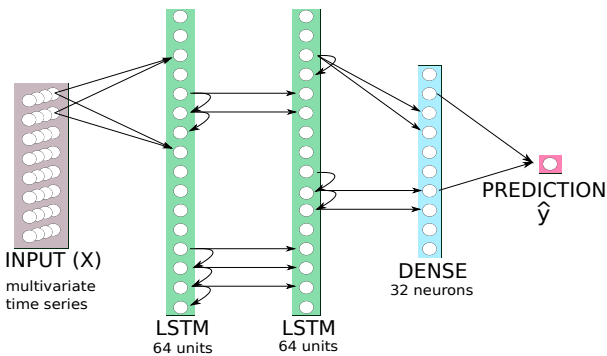}
    \caption{\label{fig:LSTM} The architecture of one constituent models of \methodname, which is an ensemble of 10 identical architectures.}
\end{figure}

\methodname\ is an ensemble of 10 LSTM-based networks, all identical in architecture, but each initialized randomly and trained separately.
Each network takes a 10-channel multivariate time series as input, has 3 hidden layers, and a single node as output.
The first two hidden layers are LSTM layers, each with 64 units.
The final hidden layer is a dense (fully-connected) layer of size 32.
The architecture is shown in Figure~\ref{fig:LSTM}.

A single LSTM unit is comprised of three `gates', which regulate the information flow through the cell and retain past information learnt from the sequence of input data.
The input gate determines new information to store from the current state, the forget gate determines what information to discard from the previous state, while the output gate determines the next hidden state or the output value of the unit.
The functioning of these gates is the reason that LSTM networks outperform `standard' RNNs in almost all tasks~\cite{Sherstinsky2020}.

\subsection{Hyperparameters}
Each individual model was trained for 15 epochs with the final model chosen as the one with the lowest training error out of the 15 epochs.
We used an Adam optimizer~\cite{Kingma2014} with a learning rate of 0.001 to minimize the mean squared error loss.
The experiments were performed in Tensorflow 2.5.0.

\subsection{\methodname\ ensemble model}
We designed \methodname\ as an ensemble model to reduce the variability of the results of a single run.
As the training of deep learning methods is a stochastic process, it may lead to different fitted models for each training; and therefore, models fit separately on the same data may generate variable results.
Ensembling generally reduces this variability, while also improving overall performance~\cite{Yang2013,Opitz1996,Fawaz2020,Lucas2019}.
Furthermore, the uncertainty in epidemiological modeling and data has resulted in ensemble models consistently outperforming individual models~\cite{Viboud2018,Johansson2019,Reich2019}.
We define our ensemble prediction as the median predicted value of all of the constituent model predictions, providing enhanced stability and consistency of our forecasts.
The curators of the COVIDhub-ensemble model investigated more sophisticated methods of combining models to form an ensemble, such as using a trained or untrained weighted mean of the constituent models, and ultimately found that taking the simple median generated equally competitive results.~\cite{Brooks2020}.

\section{Experiments}  \label{exp}
In this section, we present the results of forecasting county-level reported COVID-19 incidence using \methodname.
\rev{We compare our forecasts to 6 comparison models, including the one used by the CDC to inform decisions at a federal level, and present our results in terms of both mean absolute error and mean absolute percentage error.}
All models are assessed across 1-, 2-, 3- and 4-week forecast horizons.

\rev{After providing comparison with other leading models, we investigate the uncertainty in our forecasted values using quantile regression, and turn to look at our forecast errors and the outliers still present in the data.
Finally, we aggregate our output to investigate trends in our forecast errors at a national level.}

All experiments were run on a machine with an Intel Xenon processor and Ubuntu version 20.04.
Our models were trained using a NVIDIA GeForce RTX 3080 graphics card with 10GB of RAM.

\rev{\subsection{Comparison metrics}
We compare our forecasts to those of different models by mean average error (MAE) and mean average percentage error (MAPE).}
MAE is the metric used by the COVID-19 Forecast Hub to compare the performance of forecasts, and therefore, we have chosen to align with this decision, as the Hub is the primary reference point for COVID-19 forecasting in the US.

It is calculated as:
\begin{equation} \tag{4}
MAE_t = \frac{\sum_{j \in C}|\widehat{y_{j,t}}-y_{j,t}|}{|C|}
\end{equation}
where C is all of the counties included in the forecast, $\widehat{y_{j,t}}$ is the forecasted value for county $j$ in week $t$ and $y_{j,t}$ is the true value.

\rev{MAPE calculates the error in the forecast as a percentage of the ground truth value.
In our application, this means that the error in the forecast of each county's incidence contributes equally to the evaluation metric.
It is calculated as:
\begin{equation} \tag{5}
MAPE_t = \frac{100}{|C|} \sum_{j \in C} \frac{|\widehat{y_{j,t}}-y_{j,t}|}{y_{j,t}}
\end{equation}
where C is all of the counties included in the forecast, $\widehat{y_{j,t}}$ is the forecasted value for county $j$ in week $t$ and $y_{j,t}$ is the true value.}

\rev{As a percentage error, there are certain instances where MAPE is undefined or does not make sense~\cite{kolassa2007}.
For example, when forecasting COVID-19 at the county-level in the US, many small counties record zero weekly incidence, meaning that any forecast greater than zero will have an infinite MAPE.
For this reason, we have reported MAPE for only the 50 most populous counties in the US.}

\rev{\subsection{Comparison models} \label{comp_models}
Our experiments presented in Section~\ref{cases} compare the forecast of \methodname\ to 6 other leading forecasting models:
2 models that have been identified by the Forecast Hub as consistently high performing at the county-level~\cite{Cramer2021}; 2 models that share notable similarities to \methodname; and, 2 models produced by the Forecast Hub, one of which has been the basis of reporting and projection by the US CDC.} \\
\\
\rev{\noindent \textit{High-performing individual models:} \\
\noindent The University of Massachusetts Mechanistic Bayesian model (UMass-MechBayes)~\cite{Gibson2020}\footnote{We note that the UMass-MechBayes model only publishes forecasts for 485-490 US counties per week and the results shown in the experiments section are based on these predictions only.} and Google and Harvard University's COVID-19 Public Forecast model (Google\_Harvard-CPF)~\cite{Arik2021} are identified as consistently high performing forecasts of COVID-19 incidence at the county level in the US~\cite{Cramer2021,Ray2020}.
UMass-MechBayes is a SEIRD model modified to include non-parametric estimates of varying transmission rates and non-parametric modeling of case discrepancies to account for testing and reporting issues.
Google\_Harvard-CPF is a machine learning-infused SEIR model that emphasizes explainability.
It uses an encoder model to extract information from spatial and temporal covariates to update the transistions between compartments in the model.} \\
\\
\rev{\noindent \textit{Similar models:} \\
\noindent The Carnegie Mellon Delphi Group's Time Series model (CMU-TimeSeries)~\cite{Reinhart2021}\footnote{We note that the CMU-TimeSeries model only publishes forecasts for 199 US counties per week and the results shown in the experiments section are based on these predictions only.} and Facebook Artificial Intelligence Research's Neural Relational Autoregressive model (FAIR-NRAR)~\cite{Le2021} share similarities with our technique as they frame the problem as a data-driven forecasting problem.
CMU-TimeSeries uses incidence and deaths as inputs to an autoregressive time series model for forecasting at the county-level.
FAIR-NRAR is based on a recurrent neural network architecture and adds covariates to represent regional sociodemographics, the population mobility, and local policies.} \\
\\
\rev{\noindent \textit{Forecast Hub models:} \\
The final two models are those produced by the Forecast Hub---COVIDhub-baseline and COVIDhub-ensemble, as discussed in Section~\ref{subsec:forecasthub}.
The COVIDhub-baseline represents persistence, \ie the following week will have the same 
number of incidence as the previous week, and is developed as a universal benchmark in the US, while the COVIDhub-ensemble is the best county-level forecasting model (through ensembling multiple models created by leading universities and tech companies), and is used by the US CDC and other government departments in decision-making.}

\rev{\subsection{Forecasting COVID-19 incidence} \label{cases}
The average weekly MAE of the forecasts produced by \methodname\ and the six comparison models over the four different forecast horizons are listed in Table~\ref{tab:mae}.}

\begin{table*}[!htbp]
    \centering
    \caption{Average weekly Mean Absolute Error of COVID-19 cases per county over the whole evaluation period.}
    \label{tab:mae}
    \begin{tabular}{r|cccc}
        \toprule
        & \multicolumn{4}{c}{Forecast horizon} \\
	    & 1 week & 2 weeks & 3 weeks & 4 weeks \\
        \midrule
        CMU-TimeSeries & 810.90 & 1235.14 & 1531.42 & 1706.18 \\
        UMass-MechBayes & 457.82 & 711.11 & 962.01 & 1307.32 \\
        Google\_Harvard-CPF & 136.07 & 200.91 & 259.87 & 321.32 \\
        FAIR-NRAR & 98.31 & 156.96 & 192.08 & 213.10 \\
        COVIDhub-baseline & 91.13 & 139.55 & 180.14 & 214.62 \\
        COVIDhub-ensemble & \textbf{78.77} & 121.87 & 155.40 & 183.32 \\
        \methodname & 87.29 & \textbf{110.97} & \textbf{121.46} & \textbf{133.22} \\
        \bottomrule
    \end{tabular}
\end{table*}

\begin{figure*}[htb!]
    \centering
    \includegraphics[width=.99\linewidth]{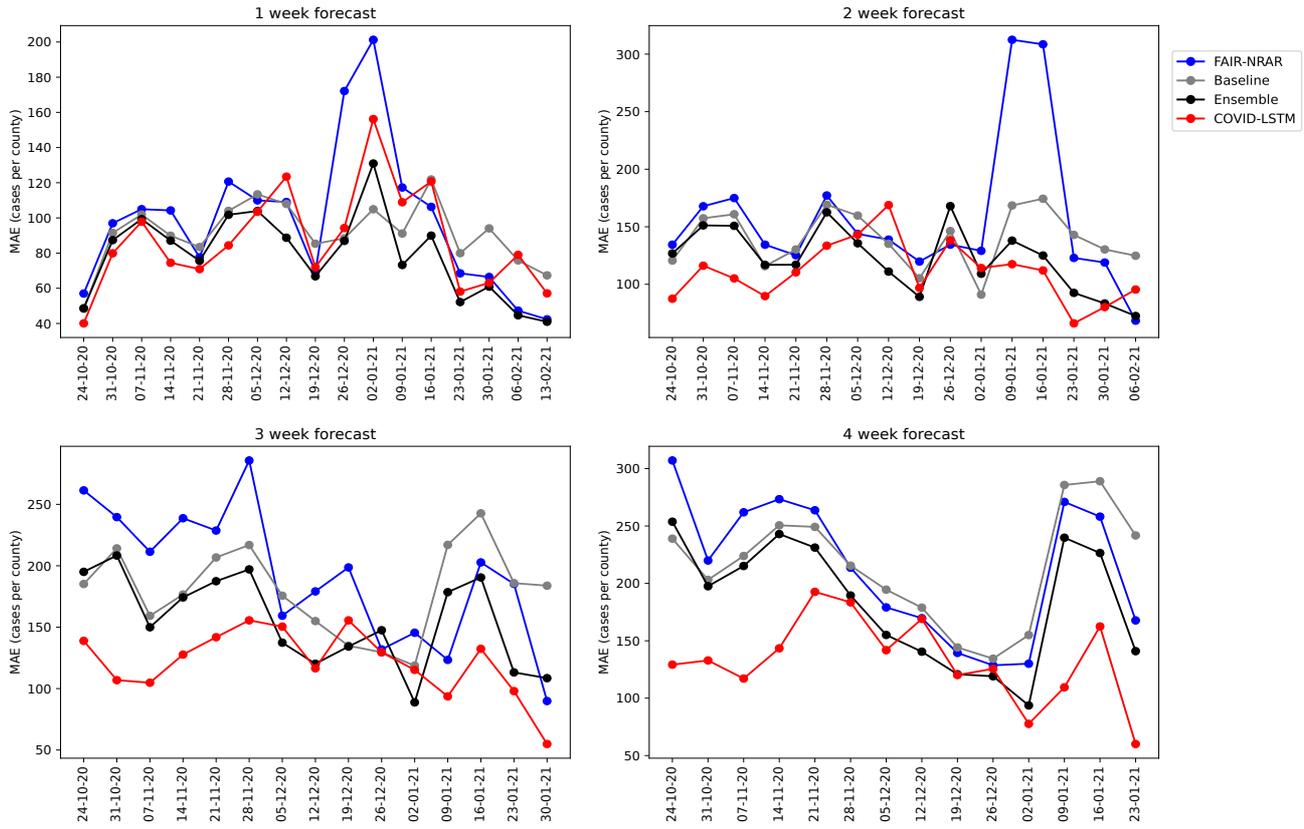}
    \caption{\label{fig:mae_plots} The mean absolute error for \methodname\ against the best published comparison models, for forecast horizons of 1 to 4 weeks in advance.}
\end{figure*}

\rev{The results show that four models---FAIR-NRAR, COVIDhub-baseline, COVIDhub-ensemble, and \methodname---have substantially lower average weekly MAE than the remaining three models.
Figure~\ref{fig:mae_plots} shows the MAE of \methodname\  against these 3 closest competitor models for each forecast date in our test period.
For the 1-week ahead forecast horizon, different models generate lower errors, depending on the specific week.
However, it is evident that \methodname\ has a lower error in many weeks across the 2-, 3-, and 4-week forecast horizons.
Over the 1-week ahead forecast, the COVIDhub-ensemble model is on average 9 cases per county more accurate than \methodname, which itself is 4 cases per county more accurate than the COVIDhub-baseline.
Over the 2-, 3-, and 4-week horizons however, \methodname\ considerably outperforms FAIR-NRAR, the COVIDhub-baseline, and COVIDhub-ensemble models on average.
Specifically, \methodname\ is on average 11, 34, and 50 cases per county more accurate than the COVIDhub-ensemble for the 2-, 3-, and 4-week horizons, respectively.
\methodname\ is also 11, 45, 71, and 80 cases per county more accurate on average than FAIR-NRAR for the 1-, 2-, 3-, and 4-week forecast horizons, respectively.
}

\rev{Figure~\ref{fig:mae_plots} shows a common trend in which our model error peaks around the national holidays in late November and late December, which may be indicative of lags and inconsistencies in reporting.
We will explore this further in Section~\ref{outliers}.}

\begin{figure*}[htb!]
    \centering
    \includegraphics[width=.99\linewidth]{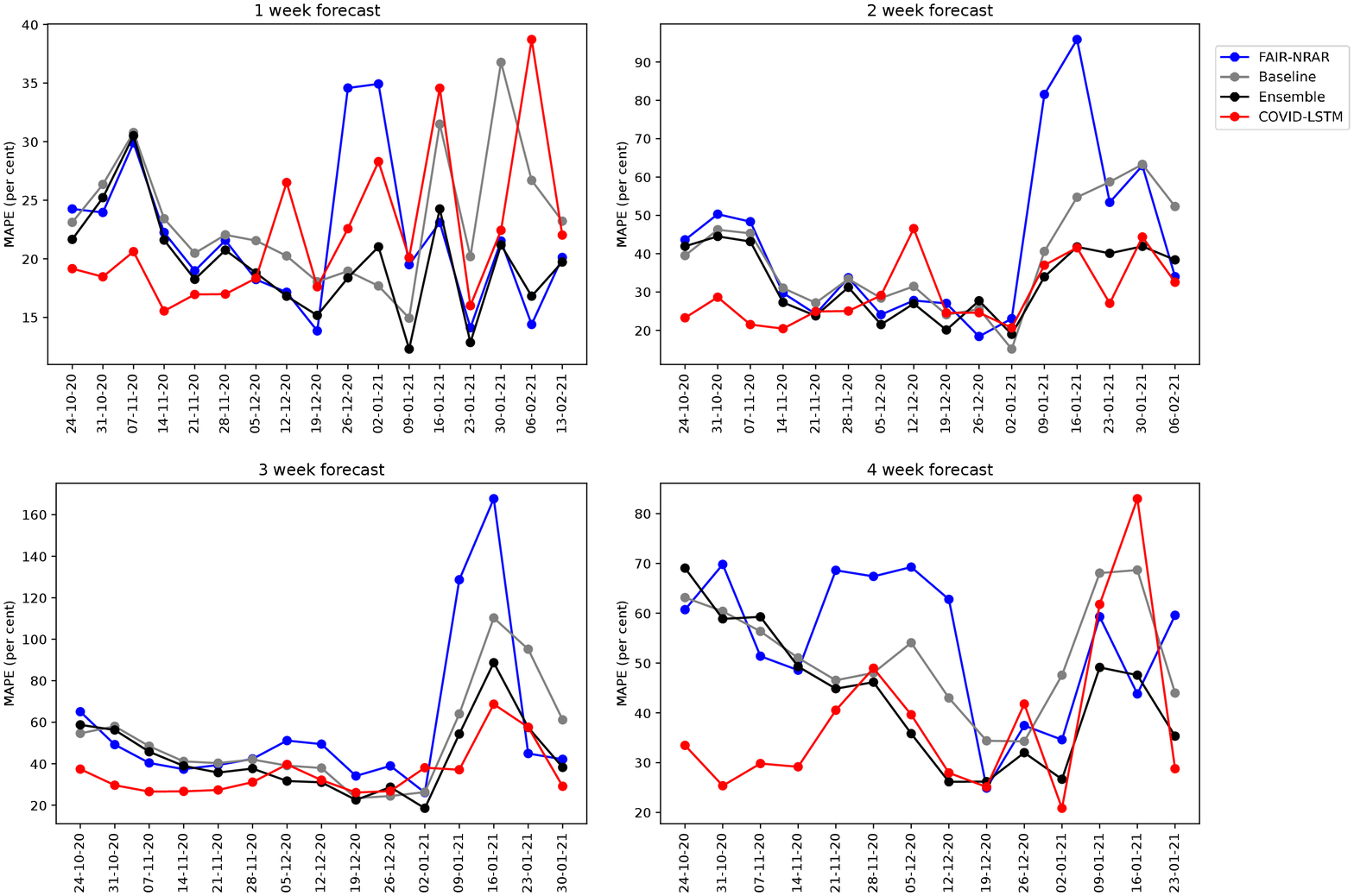}
    \caption{\label{fig:mape_plots} The mean absolute percentage error for \methodname\ against the best published comparison models, for forecast horizons of 1 to 4 weeks ahead in the 50 most populous counties in the US.}
\end{figure*}

\rev{The average weekly MAPE of the forecasts of each model is shown in Table~\ref{tab:mape}, and the MAPE of the 4 highest performing models is shown in Figure~\ref{fig:mape_plots} for the whole evaluation period.}

\begin{table*}[!htbp]
    \centering
    \caption{Average weekly Mean Absolute Percentage Error (MAPE) of COVID-19 cases per county over our evaluation period in the 50 most populous counties.}
    \label{tab:mape}
    \begin{tabular}{r|cccc}
        \toprule
        & \multicolumn{4}{c}{Forecast horizon} \\
	    & 1 week & 2 weeks & 3 weeks & 4 weeks \\
        \midrule
        CMU-TimeSeries & 24.26 & 38.01 & 49.39 & 47.31 \\
        UMass-MechBayes & 24.94 & 38.21 & 53.92 & 62.72 \\
        Google\_Harvard-CPF & 29.04 & 49.13 & 66.79 & 71.41 \\
        FAIR-NRAR & 21.91 & 42.39 & 57.12 & 54.16 \\
        COVIDhub-baseline & 23.30 & 37.64 & 51.09 & 51.39 \\
        COVIDhub-ensemble & \textbf{19.72} & 32.33 & 42.95 & 43.31 \\
        COVID-LSTM & 22.06 & \textbf{29.29} & \textbf{35.60} & \textbf{38.30} \\

        \bottomrule
    \end{tabular}
\end{table*}

\rev{The MAPE results, similar the MAE results previously reported, show that the COVIDhub-ensemble performs best at the 1-week forecast horizon while \methodname\ is best across all of the 2-, 3-, and 4-week forecast horizons.}

To further contextualize the significance of these results, in~\cite{Ray2020}, the authors stated that approximately half of the models submitted to the Forecast Hub had errors larger than the COVIDhub-baseline model.
Furthermore, the COVIDhub-ensemble model, which is used by the CDC in reporting forecasts, is an ensemble of tens of individually-calculated models each week, and the best COVID-19 Forecasting model in the US~\cite{Cramer2021}.
\methodname\ is the first COVID-19 forecasting model to outperform the COVIDhub-ensemble model on average over any forecast horizon~\cite{Cramer2021}, and it does so across two evaluation metrics and three forecast horizons.

\subsection{Prediction intervals} \label{quantiles}
\begin{figure*}[htb!]
    \centering
    \includegraphics[width=.99\linewidth]{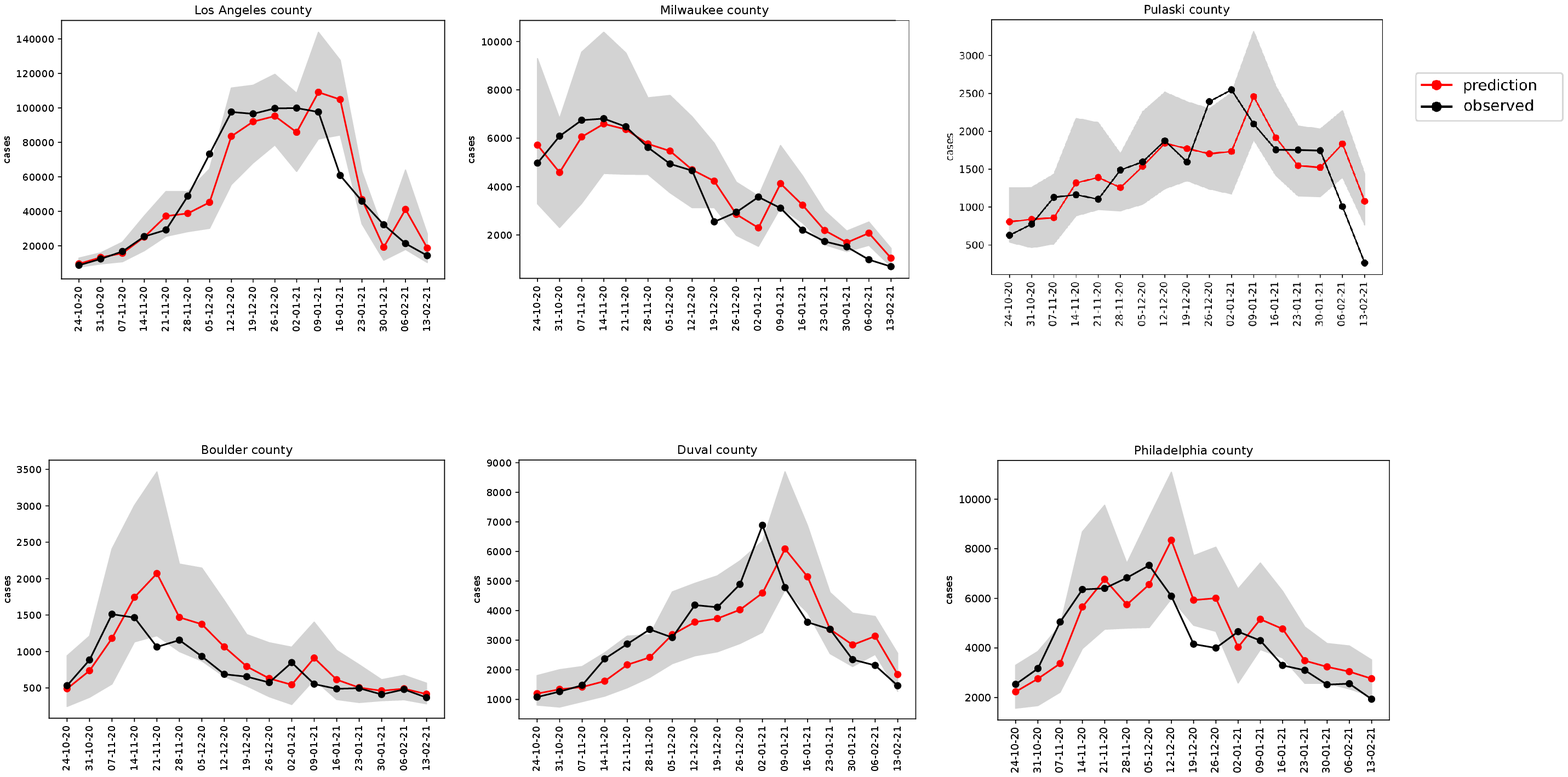}
    \caption{\label{fig:conf_int_plots} COVID-19 forecasts and 95\% prediction intervals from \methodname\ for 6 counties in the US.}
\end{figure*}

To align with the COVID-19 Forecast Hub predictions and submission requirements, \methodname\ is also capable of producing prediction intervals for each forecast.
These are produced using quantile regression adapted for deep learning~\cite{Koenker1978,Kocherginsky2005,Wei2006}.
To generate quantiles, we adapt our output layer to be of the same size as the number of quantile predictions required and modify the loss function to minimize cumulative loss over all quantiles.
Figure~\ref{fig:conf_int_plots} shows our predictions for a 1-week forecast horizon for six individual counties. The corresponding 95\% prediction interval are shown in grey.
We note that these counties--Los Angeles, California; Milwaukee, Wisconsin; Pulaski, Arkansas; Boulder, Colorado; Duval, Florida; and Philadelphia, Pennsylvania---are for illustrative purposes and have not been chosen for any specific purpose, with the exception of Los Angeles county, which has the highest number of cumulative reported COVID-19 cases in the US.

Figure~\ref{fig:conf_int_plots} shows that the large majority of our 95\% prediction intervals contain the true value.
Where the true value falls outside of the value, it is often because of a significant change in the number of reported cases that week, either an increase or a decrease.
For example, the week starting 16 January 2021 saw a decrease of approximately 40\% in recorded COVID-19 infections in Los Angeles county compared to the previous week---the week of January 9, when most likely, case numbers piled up (and under-reported) during the New Year's holidays were added to reports, and thus, was followed by a week with much fewer cases.
Likewise, and most likely for the same reason, the week starting 2 January 2021 saw an increase of approximately 40\% in Duval county compared to the previous week, and then a similar decline the following week.
In each of these examples, the reported case numbers fell outside of our model's prediction interval, but as we elaborate in the next section, this is largely due to noise in the reported data as opposed to a poor model performance.

\subsection{Outliers and rolling average incidence} \label{outliers}
The reporting of confirmed COVID-19 cases has been noisy and inconsistent in most countries worldwide; however, the inconsistencies are more pronounced at smaller scales such as US counties.
For instance, according to Johns Hopkins data~\cite{Dong2020}, 23 counties in Utah have recorded zero cases of COVID-19 since April 2020 through July 2021, including the counties of Cache, Washington, and Weber, each of which have populations exceeding 100,000 people.
Similarly, in all counties of Nebraska, zero positive cases were recorded in the months of June and July 2021.

\begin{figure}[htb!]
    \centering
    \includegraphics[width=.85\linewidth]{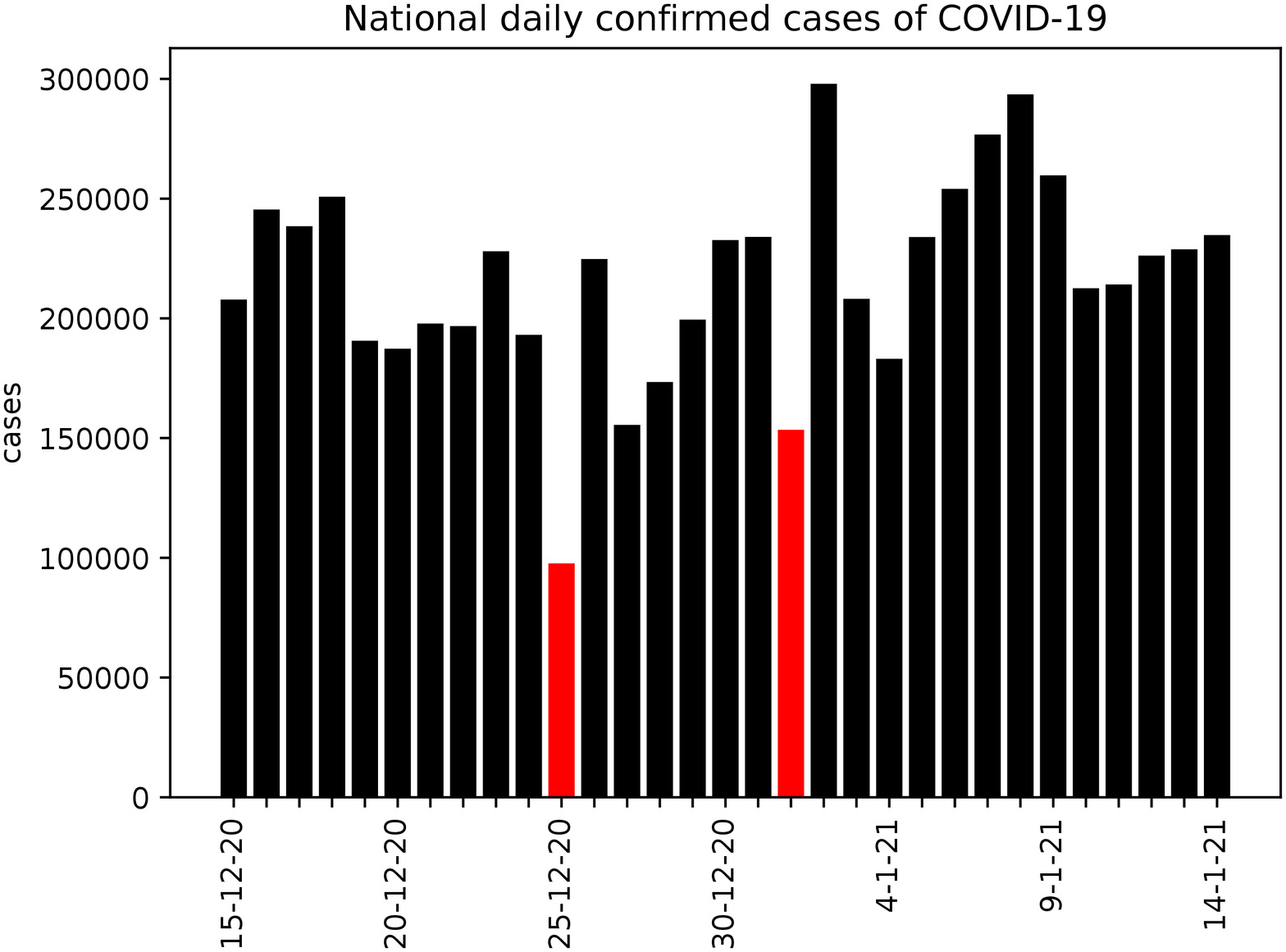}
    \caption{\label{fig:national_daily} US daily reported cases of COVID-19 highlighting the reporting variability around the 2020-21 holiday period. }
\end{figure}

Inconsistencies in county incidence reporting can also be observed at a national level, as shown in Figure~\ref{fig:national_daily}.
During the holiday period in December 2020, the national number of new daily cases dropped from 240,000 on 23 December, to below 100,000 on 25 December, and then increased again to 240,000 on December 31, only to drop by 90,000 cases the following day.
There are also observable weekly cycles in the daily data with lower cases reported on the weekends.

The variation in case data can be attributed to various causes including: the availability of tests and testers, lags in reporting, false positives, the number of asymptomatic cases, the incubation time of the virus, political motivations, and public policy~\cite{Pettengill2020,Watson2020,Surkova2020,Zhao2020,Balmford2020,Campolieti2021}.

We address the noise in the data to some extent by using a 7-day rolling average of incidence in training, instead of using the raw numbers.
This form of data smoothing is standard practice in time series analysis~\cite{Brown2004}. Although we used the smoothed values in training data, we calculated MAE presented in Figure~\ref{fig:mae_plots} against the raw data (as ground truth) without smoothing. 
It can also be argued that the smoothed data should be used as the ground truth in evaluation (\ie in calculating MAE), as otherwise the models may be penalized heavily and unfairly in weeks during which data reporting was inconsistent (\eg due to holidays or other reasons discussed above).
When we compare the forecasts of \methodname\ to the rolling average case numbers, as opposed to the raw numbers, the average MAE of our model decreases at every forecast horizon (as shown in Table~\ref{tab:mae_rolling_ave}).

\begin{figure*}[htb!]
    \centering
    \includegraphics[width=.99\linewidth]{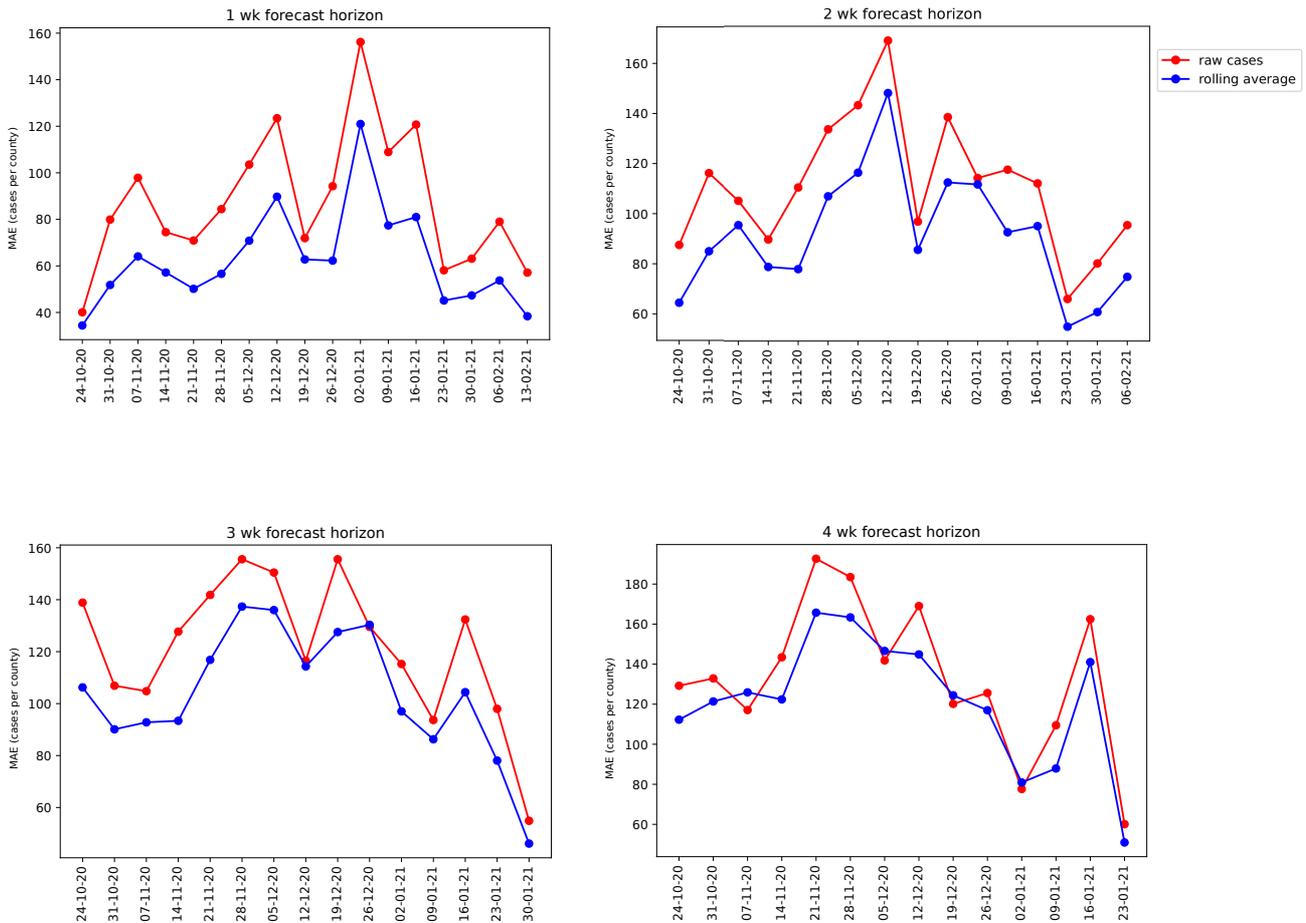}
    \caption{\label{fig:mae_rolling_ave} The mean absolute error for \methodname\ forecasts when compared to the raw confirmed cases and a 7-day rolling average of these values.}
\end{figure*}

A comparison of MAE calculated against the raw versus smoothed case numbers for the evaluation period is shown in Figure~\ref{fig:mae_rolling_ave}.
These plots show that when evaluating against the smoothed case numbers, \methodname\ performs better in almost every prediction.
While the Forecast Hub submissions are compared based on the prediction of raw case numbers, we would recommend that any decision-making or data-driven policy is based on the smoothed values of confirmed COVID-19 cases to account for inconsistencies in reporting.

\begin{table}[!htbp]
    \centering
    \caption{Comparison of the performance of \methodname\ using average weekly MAE of COVID-19 cases per county against raw case numbers and a 7-day rolling average of case numbers.}
    \label{tab:mae_rolling_ave}
    \begin{tabular}{r|cccc}
        \toprule
        & \multicolumn{4}{c}{Forecast horizon} \\
	    & 1 week & 2 weeks & 3 weeks & 4 weeks \\
        \midrule
        raw & 87.29 & 110.97 & 121.46 & 133.22 \\
        rolling ave & \textbf{62.58} & \textbf{91.33} & \textbf{103.79} & \textbf{121.76} \\
        \bottomrule
    \end{tabular}
\end{table}

\subsection{Error analysis using aggregated county-level incidence}
\begin{figure*}[htb!]
    \centering
    \includegraphics[width=.95\linewidth]{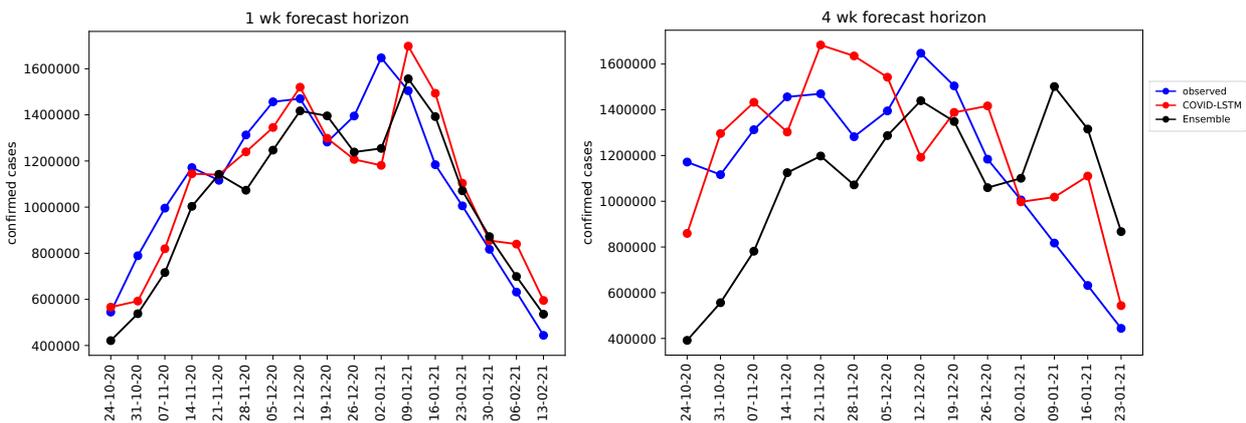}
    \caption{\label{fig:national_comparison} US weekly total reported cases of COVID-19 shown against the aggregated county-level numbers forecasted by \methodname\ and the COVIDhub-ensemble at 1- and 4-week horizons.}
\end{figure*}

Our county-level evaluations above are presented as MAE, which provides no indication if \methodname's results are consistently over- or under-estimating the true totals.
To investigate this, we have aggregated our county-level forecasts to the national-level for comparison with the the observed national incidence at both 1- and 4-week forecast horizons.
We have also aggregated the county-level forecast for the COVIDhub-ensemble for further comparison.
A plot of these are shown in Figure~\ref{fig:national_comparison}.
We note that these are not national-level forecasts as they are aimed at minimizing the per-county error, as opposed to minimizing the error in the national incidence.

The 1-week forecast horizon shows that both \methodname\ and the COVIDhub-ensemble underestimate weekly incidence when cases are rising nationally, and overestimate it when cases are declining, \ie they appear to have a bias towards persistence.
When considering the 4-week horizon, this effect is less pronounced for \methodname\ but still very evident for the COVIDhub-ensemble.
There is potential for future work to investigate this further in an attempt to the reduce bias, and improve the forecasts even further.

We note that on one forecast date---2~January~2021---both models substantially underestimate the true number of cases, which is likely due to the previous week's low data reporting during the holidays (as discussed in Section~\ref{outliers}) and the fact that this date represents the global maximum of new COVID-19 cases.

\subsection{Temporal lag} \label{lag}
\begin{figure*}[htb!]
    \centering
    \includegraphics[width=.85\linewidth]{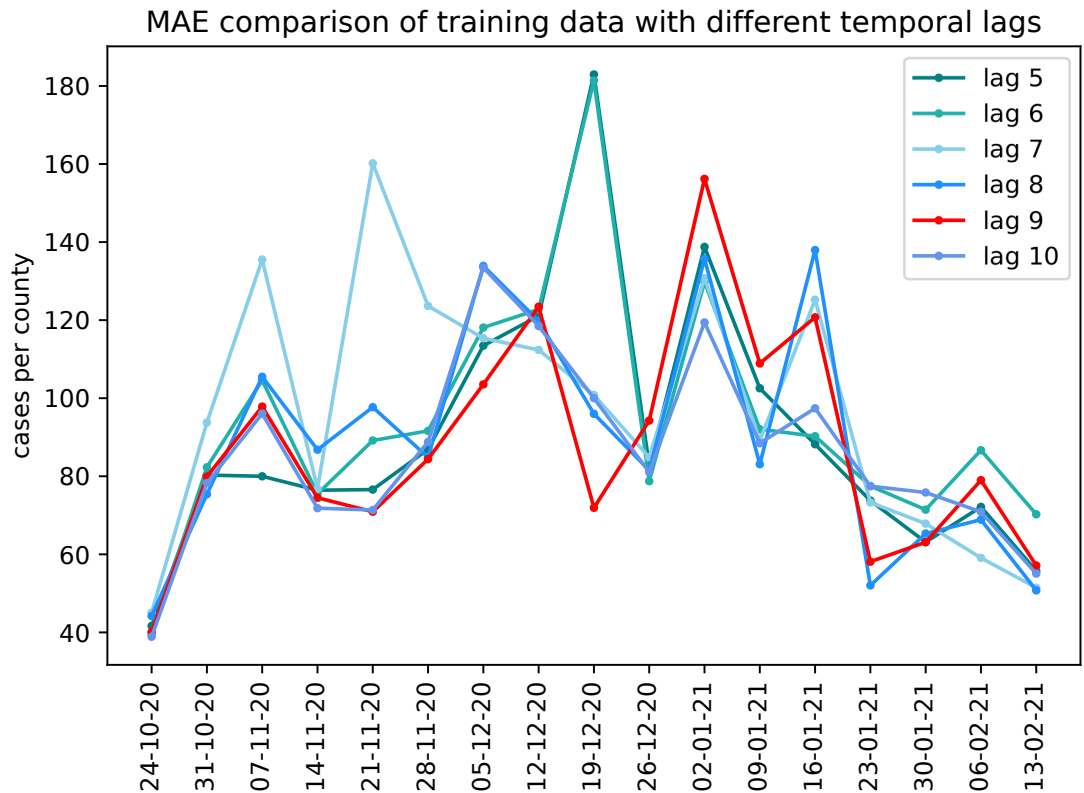}
    \caption{\label{fig:mae_lag} The mean absolute error for \methodname\ trained on input data with different temporal lags for the 17 week evaluation period. The 9-week lag (red) was chosen for use in the final model through cross-validation on the training data.}
\end{figure*}

The results presented in Section~\ref{cases} use input data with a temporal lag of 9 weeks, \ie series with a length of 10---the current week and the 9 weeks prior.
We chose this value based on an initial cross-validation performed on a smaller sample of the training data, as this is common for hyperparameter tuning in machine learning~\cite{Moore1994,Schaffer1993}.
However, there is no guarantee that the chosen value is optimal.
Figure~\ref{fig:mae_lag} shows the average MAE over the evaluation period using data with longer or shorter temporal series for a 1-week forecast horizon.
We can see that our choice to use a temporal lag of 9 weeks produces a low MAE but not conclusively the lowest.

Table~\ref{tab:mae_lag_table} shows the average MAE over the whole evaluation period.
These averages suggest that our results may have been marginally stronger if we used a lag of 10 weeks rather than 9, however this is obviously based on a comparison of evaluations on the test data, which would not be possible to see prior to performing the experiments, which are simulations of real-world model deployment.

\begin{table}[!htbp]
    \centering
    \caption{The average weekly mean absolute error for \methodname\ trained on input data with different temporal lags for the whole evaluation period}
    \label{tab:mae_lag_table}
    \begin{tabular}{c|cccccc}
        \toprule
        Temporal Lag \\ (weeks) & Average MAE \\
        \midrule
	    5 & 90.38 \\
	    6 & 94.21 \\
	    7 & 96.78 \\
	    8 & 89.38 \\
	    9 & 87.29 \\
	    10 & 86.05 \\
        \bottomrule
    \end{tabular}
\end{table}

\subsection{Socioeconomic variables}
Many models of disease spread incorporate socioeconomic and demographic variables such as median household income; proportion of the population over 65; proportion of black or Hispanic residents; and the political leanings of county residents~\cite{Garnier2021,Doti2021,Quan2021}.
Socioeconomic variables are latent variables representing factors that may cause the disease to spread faster or slower in a given county.
For example, having a lower median income may imply reduced access to healthcare or fewer white-collar workers who can work from home.

\begin{figure}[htb!]
    \centering
    \includegraphics[width=.99\linewidth]{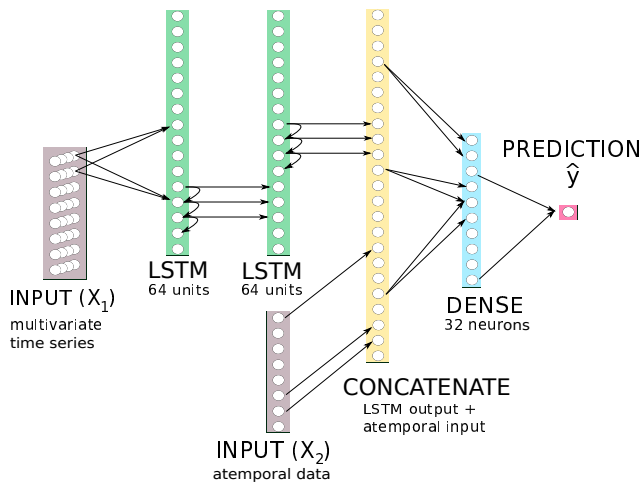}
    \caption{\label{fig:Hybrid_LSTM} The architecture of one constituent model of our Hybrid-LSTM architecture, which is an ensemble of 10 identical architectures.}
\end{figure}

In order to test whether socioeconomic and demographic variables would improve our forecasts, we defined a hybrid-LSTM model to incorporate atemporal variables alongside our existing temporal ones.
The hybrid-LSTM architecture is shown in Figure~\ref{fig:Hybrid_LSTM}.
The variables we considered in this model were: population density, proportion of black residents, proportion of Hispanic residents, proportion of indigenous residents, proportion of residents aged over 65 years, rural land as a proportion of the county area, proportion of residents who voted for Donald Trump in the 2016 US presidential election, and median household income, all at the county-level.

\begin{figure*}[htb!]
    \centering
    \includegraphics[width=.75\linewidth]{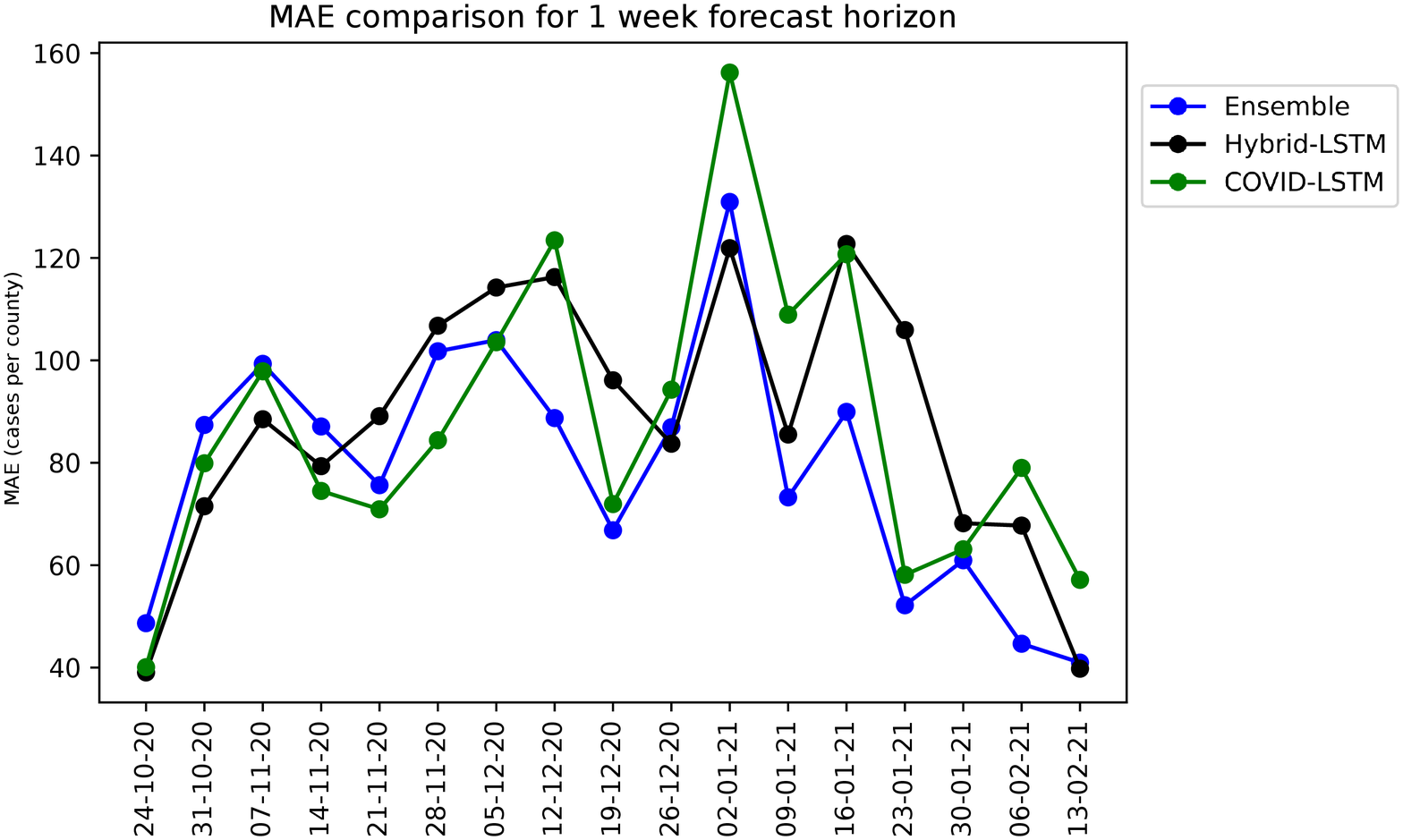}
    \caption{\label{fig:mae_hybrid} The mean absolute error for the hybrid-LSTM shown against \methodname\ and the COVIDhub-ensemble.}
\end{figure*}

A comparison of the MAE for the hybrid-LSTM model against \methodname\ and the COVIDhub-ensemble is shown in Figure~\ref{fig:mae_hybrid}.
The results show that the additional variables do not significantly improve the forecast over \methodname\ on all dates over a 1-week forecast horizon.
As listed in Table~\ref{tab:hybrid_mae}, the average MAE across the evaluation period for the hybrid-LSTM is 88.05, which is comparable with \methodname\ (87.29).
The similarity of these errors indicates that the majority of the predictive power results from the temporal features and the Facebook-derived connectedness and movement features.
These findings are also consistent with those of the Forecast Hub researchers who observed that models with fewer variables tended to perform very well in COVID-19 short-term forecasting~\cite{Cramer2021}.

\begin{table}[!htbp]
    \centering
    \caption{The mean absolute error for the hybrid-LSTM shown against \methodname,  COVIDhub-ensemble, and COVIDhub-baseline.}
    \label{tab:hybrid_mae}
    \begin{tabular}{c|c}
        \toprule
        Model & Average MAE \\
        \midrule
        COVIDhub-baseline & 91.13 \\
        COVIDhub-ensemble & 78.77 \\
        \methodname & 87.29 \\
        Hybrid-LSTM & 88.05 \\ 
        \bottomrule
    \end{tabular}
\end{table}

\section{Discussion} \label{discussion}
Our evaluation results indicate that our data-driven, spatiotemporal forecasting using deep learning is more accurate than the COVID-19 Hub Ensemble of multiple models at predicting county-level incidence over 2-, 3-, and 4-week forecast horizons.
Further, the performance metrics of our data-driven algorithm (including on the 1-week horizon) would further improve if a smoothed 7-day rolling average of incidence is used as criteria for calculating MAE.
The major benefit of a data-driven disease spread model is that the external factors that drive the spread across time and space, which are represented in the recorded case numbers for each county, are directly utilized in forecasting.
This means that we can produce highly accurate forecasts without the need to quantify variables such as the reproduction rate ($R_{0}$), which currently, are calibrated based on estimates that may not generalize to the larger population well.
We have demonstrated this by using an evaluation period that covers a time of high spread, a plateauing of the spread, and a decline in the spread of COVID-19.
However, circumventing parameters such as $R_{0}$ is a double-edged sword, as many of these variables are key indicators of disease spread, in a universally agreed-upon framework to epidemiologists.
While we have demonstrated that our data-driven method is superior to the state-of-the-art epidemiological models across most forecast horizons, the lack of these indicators may prevent its uptake in the field.
Nevertheless, we here emphasize the merit of each approach for its strength: compartmental models for characterizing parameters of a disease, and machine learning for forecasting incidence in the general population.

Another advantage of a data-driven, autoregressive, spatiotemporal model, such as \methodname, is that many variables that are difficult to measure or model are automatically captured using proxy variables.
For instance, in the US, with many local governments and states leading their own policy interventions (\eg school closures, business restrictions, masking or social distancing), keeping track of these policies and codifying them to data is next to impossible for approximately 36,000 municipalities and townships, or 13,506 school districts.
While we attempted to capture social distancing through the inclusion of daily movement ranges, we are not explicitly including any dataset listing policy mandates or business restrictions.
Further, we are not tracking mask mandates (since temporal data on masking at the county level does not exist).
However, policy intervention or individual behavior change reflects itself in the number of observed cases in the current and prior weeks, which we use in the models as an input feature.

The same can be said about the seasonality of the disease spread.
COVID-19 cases have peaked in the US both during the Fall of 2020 followed by a winter surge, as well as summer of 2021 in many parts of the country.
Our autoregressive, data-driven model captures potential seasonality by following the trend of cases in each county, based on its specific climate.

Similarly, vaccination can be a confounder in many states in late 2021.
To this date (September 2021), several states are not making county-level data available for vaccination, and the rate of vaccination has been spatially heterogeneous across the country.
While our model evaluation period presented in this paper does not overlap with the public availability of vaccines, in general, more vaccinations in a county would lead to a drop to new infections, which would be captured by our autoregressive model on-the-fly, as opposed to the traditional compartmental models that may require updating of the model parameters and incorporating new assumptions.
We leave the incorporation of vaccination statistics and investigating its potential effect on forecasting performance for future research.

It is worth noting that while Facebook-derived datasets may not be perfectly representative of different demographic layers in the US, our models generate county-level forecasts (in a similar fashion used by the Forecast Hub and the CDC), and not age-, race-, or gender-specific forecasts.
Nevertheless, in terms of representation, Facebook has more than 200 million users in the US, resulting in one of the largest datasets available on human movement and county-connectivity.
Furthermore, Facebook, which has 2.89 billion users worldwide, releases similar datasets for other countries as well, including locations where other sources of data are scarce.

We also note that our data-driven models benefit from the wealth of data present for COVID-19 cases when compared to previous pandemics or epidemics.
If the data were more sparse---\ie. more values of zero recorded---a LSTM-based model would have more difficulty learning patterns~\cite{Liu2021,Drumond2018}.

Based on the information presented in this section and the overall accuracy of \methodname, we argue that disease spread modeling should move to a realm of coexistence of compartmental models and data-driven models.

\rev{\subsection{Future opportunities}
While our model produces highly accurate forecasts, we acknowledge that there are outstanding challenges, which if addressed, may further improve its performance, and the performance of COVID-19 forecasting overall.}

\rev{First, while we have accounted for some of the noise in the data through smoothing, a systematic algorithm for identifying and handling irregularities in the county-level data would likely improve model performance. Researchers at the Forecast Hub in the U.S. have invested significant effort for building automated anomaly detection methods; however, they report that their experiments did not yield consistently satisfactory results that improve over human judgment~\cite{Reich2020}.}

\rev{Secondly, while our model can account for vaccination rates through weekly incidence, an explicit input variable representing the proportion of vaccinated residents in the county may improve the model further.
We note that this is not publicly available at the county-level in the US at the time of writing (September 2021).
There is also scope to explore the relationship between movement variables and vaccination status, \ie whether the people moving about are or are not vaccinated.}

\rev{Thirdly, forecasts will improve from a deeper understanding of the disease characteristics.
Studies have found different values for $R_0$ for different geographical areas and different stages of the pandemic, ranging from 3.1 in Brazil~\cite{deSouza2020} to 5.7 in Wuhan, China~\cite{Sanche2020}.
Similarly, the incubation period is widely reported as 5 days; however, it has been measured as potentially being up to 14 days and likely depends on the variant~\cite{Lauer2020}.
An increase in the genomic sequencing of confirmed COVID-19 cases, and further understanding of the transmissibility of different variants of the disease, may also help improve forecasting performance. However, at present, very little genomic sequencing of confirmed cases in the US is being conducted compared to other countries~\cite{Furuse2021,Robishaw2021}.}

\rev{While the above represent notable challenges and opportunities for improving future COVID-19 forecasts, there are many other complexities of the data and disease (as identified in~\cite{Cao2021}), which a deeper understanding of will lead to improved forecasting performance.}

\section{Conclusion} \label{conclusion}
In this paper, we presented \methodname, a data-driven approach to county-level forecasting of COVID-19.
We approached the task as a spatiotemporal machine learning problem by using a temporal series of cases and hand-engineered spatial features derived from Facebook movement and connectedness datasets.
\methodname\ outperforms the COVIDhub-ensemble on our 17-week evaluation period, making it the first model to be more accurate than the COVIDhub-ensemble over one or more forecast periods.
Specifically, over the 2-, 3-, and 4-week forecast horizons, \methodname\ is 11, 34, and 50 cases per county more accurate than the COVIDhub-ensemble.

The high predictive power of our deep learning-based approach for forecasting COVID-19 incidence at high spatial resolutions is notable, especially given its incorporation of spatial features derived from Facebook.
Facebook has over 2.89 users worldwide, and releases similar datasets for many other countries.
This means that a similar forecasting approach could be useful in data-poor regions or where other sources of data may be scarce.

We have demonstrated that a data-driven spatiotemporal approach to forecasting would be greatly informative and beneficial for decision-making and planning purposes.
We also acknowledge that 1) our conclusions are valid at the spatial scale and for the study area/period that we have used, not necessarily for any specific region, gender, or age group, and 2)
forecasts are not the only output of compartmental epidemiological models, and that the uptake of data-driven approaches in the field may require future work to integrate and utilize both data-driven and traditional epidemiological models.

Future research should investigate methods for calibrating compartmental models in the framework of deep learning, to offer the benefits of both models: more precise forecasting, and better characterization of the contagion. Another worthwhile research direction includes designing deep architectures for direct incorporation of spatial features, to potentially improve upon our current approach of hand-engineered spatial features. Lastly, future research should investigate the incorporation of vaccination variables for forecasting at high spatial resolutions, especially that annual booster shots and seasonal COVID-19 epidemics may define the world's new norm.

\section*{Acknowledgements}
This work was supported by the Population Council, and the University of Colorado Population Center (CUPC) funded by Eunice Kennedy Shriver National Institute of Child Health \& Human Development of the National Institutes of Health (P2CHD066613). The content is solely the responsibility of the authors, and does not reflect the views of the Population Council, or official views of the NIH, CUPC, or the University of Colorado.

\section*{Conflicts of interest}
The authors have no relevant financial or non-financial interests to disclose.

\section*{Supplementary material}
To aid replication, the code for our method and the raw results of all experiments are available at \repo{}.


\begin{thebibliography}{100}

\providecommand{\url}[1]{{#1}}
\providecommand{\urlprefix}{URL }
\expandafter\ifx\csname urlstyle\endcsname\relax
  \providecommand{\doi}[1]{DOI~\discretionary{}{}{}#1}\else
  \providecommand{\doi}{DOI~\discretionary{}{}{}\begingroup
  \urlstyle{rm}\Url}\fi

\bibitem{Ansumali2020}
Ansumali, S., Prakash, M.K.: A very flat peak: Why standard {SEIR} models miss
  the plateau of {COVID-19} infections and how it can be corrected.
\newblock medRxiv  (2020)

\bibitem{Arik2021}
Arik, S.O., Li, C.L., Yoon, J., Sinha, R., Epshteyn, A., Le, L.T., Menon, V.,
  Singh, S., Zhang, L., Yoder, N., Nikoltchev, M., Sonthalia, Y., Nakhost, H.,
  Kanal, E., Pfister, T.: Interpretable sequence learning for {COVID-19}
  forecasting.
\newblock arXiv preprint \textbf{2008.00646} (2021).
\newblock \urlprefix\url{http://arxiv.org/abs/2008.00646}

\bibitem{Astin2021}
Astin~Cole, H.A., Ahmed, A., Hamasha, M., Jordan, S.: Identifying patterns of
  turnover intention among alabama frontline nurses in hospital settings during
  the {COVID-19} pandemic.
\newblock Journal of Multidisciplinary Healthcare \textbf{14}, 1783 (2021)

\bibitem{Bailey2018}
Bailey, M., Cao, R., Kuchler, T., Stroebel, J., Wong, A.: Social connectedness:
  Measurement, determinants, and effects.
\newblock Journal of Economic Perspectives \textbf{32}(3), 259--80 (2018)

\bibitem{Balmford2020}
Balmford, B., Annan, J.D., Hargreaves, J.C., Alto{\`e}, M., Bateman, I.J.:
  Cross-country comparisons of covid-19: policy, politics and the price of
  life.
\newblock Environmental and Resource Economics \textbf{76}(4), 525--551 (2020)

\bibitem{Bandyopadhyay2020}
Bandyopadhyay, S.K., Dutta, S.: Machine learning approach for confirmation of
  {COVID-19} cases: Positive, negative, death and release.
\newblock MedRxiv  (2020)

\bibitem{Bashir2020}
Bashir, M.F., Ma, B., Bilal, Komal, B., Bashir, M.A., Tan, D., Bashir, M.:
  Correlation between climate indicators and {COVID-19} pandemic in {New York,
  USA}.
\newblock Science of The Total Environment \textbf{728}, 138835 (2020).
\newblock \doi{https://doi.org/10.1016/j.scitotenv.2020.138835}

\bibitem{Beck2020}
Beck, B.R., Shin, B., Choi, Y., Park, S., Kang, K.: Predicting commercially
  available antiviral drugs that may act on the novel coronavirus
  ({SARS-CoV-2}) through a drug-target interaction deep learning model.
\newblock Computational and structural biotechnology journal \textbf{18},
  784--790 (2020)

\bibitem{Brinati2020}
Brinati, D., Campagner, A., Ferrari, D., Locatelli, M., Banfi, G., Cabitza, F.:
  Detection of {COVID-19} infection from routine blood exams with machine
  learning: a feasibility study.
\newblock Journal of medical systems \textbf{44}(8), 1--12 (2020)

\bibitem{Brooks2020}
Brooks, L.C., Ray, E.L., Bien, J., Bracher, J., Rumack, A., Tibshirani, R.J.,
  Reich, N.G.: Comparing ensemble approaches for short-term probabilistic
  {COVID-19} forecasts in the {US}.
\newblock International Institute of Forecasters  (2020)

\bibitem{Brown2004}
Brown, R.G.: Smoothing, forecasting and prediction of discrete time series.
\newblock Courier Corporation (2004)

\bibitem{Campolieti2021}
Campolieti, M.: {COVID-19} deaths in the {USA}: Benford’s law and
  under-reporting.
\newblock Journal of Public Health (Oxford, England)  (2021)

\bibitem{Cao2021}
Cao, L., Liu, Q.: Covid-19 modeling: A review.
\newblock arXiv preprint \textbf{2104.12556} (2021).
\newblock \urlprefix\url{http://arxiv.org/abs/2104.12556}

\bibitem{Castillo1989}
Castillo-Chavez, C., Hethcote, H.W., Andreasen, V., Levin, S.A., Liu, W.M.:
  Epidemiological models with age structure, proportionate mixing, and
  cross-immunity.
\newblock Journal of mathematical biology \textbf{27}(3), 233--258 (1989)

\bibitem{Chimmula2020}
Chimmula, V.K.R., Zhang, L.: Time series forecasting of {COVID-19} transmission
  in canada using {LSTM} networks.
\newblock Chaos, Solitons \& Fractals \textbf{135}, 109864 (2020)

\bibitem{Chinazzi2020}
Chinazzi, M., Davis, J.T., Ajelli, M., Gioannini, C., Litvinova, M., Merler,
  S., y~Piontti, A.P., Mu, K., Rossi, L., Sun, K., et~al.: The effect of travel
  restrictions on the spread of the 2019 novel coronavirus (covid-19) outbreak.
\newblock Science \textbf{368}(6489), 395--400 (2020)

\bibitem{Chu2021}
Chu, J., Ghenand, O., Collins, J., Byrne, J., Wentworth, A., Chai, P.R.,
  Dadabhoy, F., Hur, C., Traverso, G.: Thinking green: modelling respirator
  reuse strategies to reduce cost and waste.
\newblock BMJ Open \textbf{11}(7) (2021).
\newblock \doi{10.1136/bmjopen-2021-048687}

\bibitem{Cohen2020}
Cohen, J., Rodgers, Y.v.d.M.: Contributing factors to personal protective
  equipment shortages during the {COVID}-19 pandemic.
\newblock Preventive medicine \textbf{141}, 106263 (2020)

\bibitem{Cramer2021}
Cramer, E.Y., Ray, E.L., Lopez, V.K., Bracher, J., Brennen, A., Rivadeneira,
  A.J.C., Gerding, A., Gneiting, T., House, K.H., Huang, Y., Jayawardena, D.,
  Kanji, A.H., Khandelwal, A., Le, K., M{\"u}hlemann, A., Niemi, J., Shah, A.,
  Stark, A., Wang, Y., Wattanachit, N., Zorn, M.W., Gu, Y., Jain, S., Bannur,
  N., Deva, A., Kulkarni, M., Merugu, S., Raval, A., Shingi, S., Tiwari, A.,
  White, J., Woody, S., Dahan, M., Fox, S., Gaither, K., Lachmann, M., Meyers,
  L.A., Scott, J.G., Tec, M., Srivastava, A., George, G.E., Cegan, J.C.,
  Dettwiller, I.D., England, W.P., Farthing, M.W., Hunter, R.H., Lafferty, B.,
  Linkov, I., Mayo, M.L., Parno, M.D., Rowland, M.A., Trump, B.D., Corsetti,
  S.M., Baer, T.M., Eisenberg, M.C., Falb, K., Huang, Y., Martin, E.T.,
  McCauley, E., Myers, R.L., Schwarz, T., Sheldon, D., Gibson, G.C., Yu, R.,
  Gao, L., Ma, Y., Wu, D., Yan, X., Jin, X., Wang, Y.X., Chen, Y., Guo, L.,
  Zhao, Y., Gu, Q., Chen, J., Wang, L., Xu, P., Zhang, W., Zou, D., Biegel, H.,
  Lega, J., Snyder, T.L., Wilson, D.D., McConnell, S., Walraven, R., Shi, Y.,
  Ban, X., Hong, Q.J., Kong, S., Turtle, J.A., Ben-Nun, M., Riley, P., Riley,
  S., Koyluoglu, U., DesRoches, D., Hamory, B., Kyriakides, C., Leis, H.,
  Milliken, J., Moloney, M., Morgan, J., Ozcan, G., Schrader, C., Shakhnovich,
  E., Siegel, D., Spatz, R., Stiefeling, C., Wilkinson, B., Wong, A., Gao, Z.,
  Bian, J., Cao, W., Ferres, J.L., Li, C., Liu, T.Y., Xie, X., Zhang, S.,
  Zheng, S., Vespignani, A., Chinazzi, M., Davis, J.T., Mu, K., Piontti,
  A.P.y., Xiong, X., Zheng, A., Baek, J., Farias, V., Georgescu, A., Levi, R.,
  Sinha, D., Wilde, J., Penna, N.D., Celi, L.A., Sundar, S., Cavany, S.,
  Espa{\~n}a, G., Moore, S., Oidtman, R., Perkins, A., Osthus, D., Castro, L.,
  Fairchild, G., Michaud, I., Karlen, D., Lee, E.C., Dent, J., Grantz, K.H.,
  Kaminsky, J., Kaminsky, K., Keegan, L.T., Lauer, S.A., Lemaitre, J.C.,
  Lessler, J., Meredith, H.R., Perez-Saez, J., Shah, S., Smith, C.P., Truelove,
  S.A., Wills, J., Kinsey, M., Obrecht, R., Tallaksen, K., Burant, J.C., Wang,
  L., Gao, L., Gu, Z., Kim, M., Li, X., Wang, G., Wang, Y., Yu, S., Reiner,
  R.C., Barber, R., Gaikedu, E., Hay, S., Lim, S., Murray, C., Pigott, D.,
  Prakash, B.A., Adhikari, B., Cui, J., Rodr{\'\i}guez, A., Tabassum, A., Xie,
  J., Keskinocak, P., Asplund, J., Baxter, A., Oruc, B.E., Serban, N., Arik,
  S.O., Dusenberry, M., Epshteyn, A., Kanal, E., Le, L.T., Li, C.L., Pfister,
  T., Sava, D., Sinha, R., Tsai, T., Yoder, N., Yoon, J., Zhang, L., Abbott,
  S., Bosse, N.I., Funk, S., Hellewel, J., Meakin, S.R., Munday, J.D.,
  Sherratt, K., Zhou, M., Kalantari, R., Yamana, T.K., Pei, S., Shaman, J.,
  Ayer, T., Adee, M., Chhatwal, J., Dalgic, O.O., Ladd, M.A., Linas, B.P.,
  Mueller, P., Xiao, J., Li, M.L., Bertsimas, D., Lami, O.S., Soni, S.,
  Bouardi, H.T., Wang, Y., Wang, Q., Xie, S., Zeng, D., Green, A., Bien, J.,
  Hu, A.J., Jahja, M., Narasimhan, B., Rajanala, S., Rumack, A., Simon, N.,
  Tibshirani, R., Tibshirani, R., Ventura, V., Wasserman, L.,
  O{\textquoteright}Dea, E.B., Drake, J.M., Pagano, R., Walker, J.W., Slayton,
  R.B., Johansson, M., Biggerstaff, M., Reich, N.G.: Evaluation of individual
  and ensemble probabilistic forecasts of {COVID}-19 mortality in the {U.S.}
\newblock medRxiv  (2021).
\newblock \doi{10.1101/2021.02.03.21250974}

\bibitem{Das2020}
Das, D., Santosh, K., Pal, U.: Truncated inception net: Covid-19 outbreak
  screening using chest x-rays.
\newblock Physical and engineering sciences in medicine \textbf{43}(3),
  915--925 (2020)

\bibitem{Dean2020}
Dean, N.E., y~Piontti, A.P., Madewell, Z.J., Cummings, D.A., Hitchings, M.D.,
  Joshi, K., Kahn, R., Vespignani, A., Halloran, M.E., Longini~Jr, I.M.:
  Ensemble forecast modeling for the design of {COVID-19} vaccine efficacy
  trials.
\newblock Vaccine \textbf{38}(46), 7213--7216 (2020)

\bibitem{Dehning2020}
Dehning, J., Zierenberg, J., Spitzner, F.P., Wibral, M., Neto, J.P., Wilczek,
  M., Priesemann, V.: Inferring change points in the spread of {COVID-19}
  reveals the effectiveness of interventions.
\newblock Science \textbf{369}(6500) (2020)

\bibitem{Dong2020}
Dong, E., Du, H., Gardner, L.: An interactive web-based dashboard to track
  {COVID-19} in real time.
\newblock The Lancet Infectious Diseases \textbf{20}(5), 533--534 (2020)

\bibitem{Doti2021}
Doti, J.L.: Examining the impact of socioeconomic variables on {COVID-19} death
  rates at the state level.
\newblock Journal of Bioeconomics \textbf{23}(1), 15--53 (2021)

\bibitem{Drumond2018}
Drumond, R.R., Marques, B.A.D., Vasconcelos, C.N., Clua, E.: An {LSTM}
  recurrent network for motion classification from sparse data.
\newblock In: Proceedings of the 13th International Joint Conference on
  Computer Vision, Imaging and Computer Graphics Theory and Applications,
  vol.~1, pp. 215--22 (2018)

\bibitem{Fang2020}
Fang, Y., Nie, Y., Penny, M.: Transmission dynamics of the covid-19 outbreak
  and effectiveness of government interventions: A data-driven analysis.
\newblock Journal of medical virology \textbf{92}(6), 645--659 (2020)

\bibitem{Fawaz2020}
Fawaz, H.I., Lucas, B., Forestier, G., Pelletier, C., Schmidt, D.F., Weber, J.,
  Webb, G.I., Idoumghar, L., Muller, P.A., Petitjean, F.: Inceptiontime:
  Finding alexnet for time series classification.
\newblock Data Mining and Knowledge Discovery \textbf{34}(6), 1936--1962 (2020)

\bibitem{Flaxman2020}
Flaxman, S., Mishra, S., Gandy, A., Unwin, H.J.T., Mellan, T.A., Coupland, H.,
  Whittaker, C., Zhu, H., Berah, T., Eaton, J.W., et~al.: Estimating the
  effects of non-pharmaceutical interventions on {COVID-19} in europe.
\newblock Nature \textbf{584}(7820), 257--261 (2020)

\bibitem{Friedman2021}
Friedman, J., Liu, P., Troeger, C.E., Carter, A., Reiner, R.C., Barber, R.M.,
  Collins, J., Lim, S.S., Pigott, D.M., Vos, T., et~al.: Predictive performance
  of international {COVID-19} mortality forecasting models.
\newblock Nature communications \textbf{12}(1), 1--13 (2021)

\bibitem{Furuse2021}
Furuse, Y.: Genomic sequencing effort for sars-cov-2 by country during the
  pandemic.
\newblock International Journal of Infectious Diseases \textbf{103}, 305--307
  (2021)

\bibitem{Galanis2021}
Galanis, P., Vraka, I., Fragkou, D., Bilali, A., Kaitelidou, D.: Nurses'
  burnout and associated risk factors during the {COVID-19} pandemic: A
  systematic review and meta-analysis.
\newblock Journal of advanced nursing  (2021)

\bibitem{Garnier2021}
Garnier, R., Benetka, J.R., Kraemer, J., Bansal, S.: Socioeconomic disparities
  in social distancing during the {COVID-19} pandemic in the {United States}:
  observational study.
\newblock Journal of medical Internet research \textbf{23}(1), e24591 (2021)

\bibitem{Gautam2021}
Gautam, Y.: Transfer learning for {COVID-19} cases and deaths forecast using
  {LSTM} network.
\newblock ISA transactions  (2021)

\bibitem{Gers2000}
Gers, F.A., Schmidhuber, J., Cummins, F.: {Learning to Forget: Continual
  Prediction with {LSTM}}.
\newblock Neural Computation \textbf{12}(10), 2451--2471 (2000).
\newblock \doi{10.1162/089976600300015015}

\bibitem{Getz2019}
Getz, W.M., Salter, R., Mgbara, W.: Adequacy of {SEIR} models when epidemics
  have spatial structure: {E}bola in {S}ierra {L}eone.
\newblock Philosophical Transactions of the Royal Society \textbf{374}(1775),
  20180282 (2019)

\bibitem{Gibson2020}
Gibson, G.C., Reich, N.G., Sheldon, D.: Real-time mechanistic bayesian
  forecasts of covid-19 mortality.
\newblock medRxiv  (2020).
\newblock
  \urlprefix\url{https://www.medrxiv.org/content/early/2020/12/24/2020.12.22.20248736}

\bibitem{Graves2008}
Graves, A., Liwicki, M., Fernández, S., Bertolami, R., Bunke, H., Schmidhuber,
  J.: A novel connectionist system for unconstrained handwriting recognition.
\newblock IEEE Transactions on Pattern Analysis and Machine Intelligence
  \textbf{31}(5), 855--868 (2009).
\newblock \doi{10.1109/TPAMI.2008.137}

\bibitem{Halevy2009}
Halevy, A., Norvig, P., Pereira, F.: The unreasonable effectiveness of data.
\newblock IEEE Intelligent Systems \textbf{24}(2), 8--12 (2009)

\bibitem{He2020}
He, S., Peng, Y., Sun, K.: {SEIR} modeling of the {COVID-19} and its dynamics.
\newblock Nonlinear dynamics \textbf{101}(3), 1667--1680 (2020)

\bibitem{herdagdelen2020}
Herdağdelen, A., Dow, A.: Protecting privacy in facebook mobility data during
  the {COVID-19} response (2020).
\newblock
  \urlprefix\url{https://research.fb.com/blog/2020/06/protecting-privacy-in-facebook-mobility-data-during-the-covid-19-response/}

\bibitem{Hethcote2000}
Hethcote, H.W.: The mathematics of infectious diseases.
\newblock SIAM review \textbf{42}(4), 599--653 (2000)

\bibitem{Hochreiter1997}
Hochreiter, S., Schmidhuber, J.: {Long Short-Term Memory}.
\newblock Neural Computation \textbf{9}(8), 1735--1780 (1997).
\newblock \doi{10.1162/neco.1997.9.8.1735}

\bibitem{Hong2020}
Hong, H.G., Li, Y.: Estimation of time-varying reproduction numbers underlying
  epidemiological processes: A new statistical tool for the covid-19 pandemic.
\newblock PLOS ONE \textbf{15}(7), 1--15 (2020)

\bibitem{Ibrahim2021}
Ibrahim, M.R., Haworth, J., Lipani, A., Aslam, N., Cheng, T., Christie, N.:
  Variational-{LSTM} autoencoder to forecast the spread of coronavirus across
  the globe.
\newblock PloS one \textbf{16}(1), e0246120 (2021)

\bibitem{IHME2020}
{IHME COVID-19 forecasting team}: Modeling {COVID-19} scenarios for the
  {U}nited {S}tates.
\newblock Nature medicine  (2020)

\bibitem{Johansson2019}
Johansson, M.A., Apfeldorf, K.M., Dobson, S., Devita, J., Buczak, A.L.,
  Baugher, B., Moniz, L.J., Bagley, T., Babin, S.M., Guven, E., et~al.: An open
  challenge to advance probabilistic forecasting for dengue epidemics.
\newblock Proceedings of the National Academy of Sciences \textbf{116}(48),
  24268--24274 (2019)

\bibitem{Keeling2020}
Keeling, M.J., Hollingsworth, T.D., Read, J.M.: Efficacy of contact tracing for
  the containment of the 2019 novel coronavirus ({COVID-19}).
\newblock J Epidemiol Community Health \textbf{74}(10), 861--866 (2020)

\bibitem{Kendall1956}
Kendall, D.G.: {D}eterministic and stochastic epidemics in closed populations,
  pp. 149--166.
\newblock University of California Press (1956).
\newblock \doi{doi:10.1525/9780520350717-011}

\bibitem{Kingma2014}
Kingma, D.P., Ba, J.: Adam: A method for stochastic optimization.
\newblock arXiv preprint \textbf{1412.6980} (2014).
\newblock \urlprefix\url{http://arxiv.org/abs/1412.6980}

\bibitem{Kocherginsky2005}
Kocherginsky, M., He, X., Mu, Y.: Practical confidence intervals for regression
  quantiles.
\newblock Journal of Computational and Graphical Statistics \textbf{14}(1),
  41--55 (2005).
\newblock \doi{10.1198/106186005X27563}

\bibitem{Koelle2006}
Koelle, K., Cobey, S., Grenfell, B., Pascual, M.: Epochal evolution shapes the
  phylodynamics of interpandemic influenza a (h3n2) in humans.
\newblock Science \textbf{314}(5807), 1898--1903 (2006)

\bibitem{Koenker1978}
Koenker, R., Bassett, G.: Regression quantiles.
\newblock Econometrica \textbf{46}(1), 33--50 (1978)

\bibitem{Koh2020}
Koh, D.: {COVID-19} lockdowns throughout the world.
\newblock Occupational Medicine \textbf{70}(5), 322--322 (2020)

\bibitem{kolassa2007}
Kolassa, S., Sch{\"u}tz, W., et~al.: Advantages of the mad/mean ratio over the
  mape.
\newblock Foresight: The International Journal of Applied Forecasting
  \textbf{6}, 40--43 (2007)

\bibitem{Kontis2020}
Kontis, V., Bennett, J.E., Rashid, T., Parks, R.M., Pearson-Stuttard, J.,
  Guillot, M., Asaria, P., Zhou, B., Battaglini, M., Corsetti, G., et~al.:
  Magnitude, demographics and dynamics of the effect of the first wave of the
  covid-19 pandemic on all-cause mortality in 21 industrialized countries.
\newblock Nature medicine \textbf{26}(12), 1919--1928 (2020)

\bibitem{Kuchler2021}
Kuchler, T., Russel, D., Stroebel, J.: The geographic spread of {COVID-19}
  correlates with the structure of social networks as measured by facebook.
\newblock Journal of Urban Economics p. 103314 (2021)

\bibitem{Lauer2020}
Lauer, S.A., Grantz, K.H., Bi, Q., Jones, F.K., Zheng, Q., Meredith, H.R.,
  Azman, A.S., Reich, N.G., Lessler, J.: The incubation period of coronavirus
  disease 2019 {(COVID-19)} from publicly reported confirmed cases: estimation
  and application.
\newblock Annals of internal medicine \textbf{172}(9), 577--582 (2020)

\bibitem{Le2021}
Le, M., Ibrahim, M., Sagun, L., Lacroix, T., Nickel, M.: Neural relational
  autoregression for high-resolution {COVID-19} forecasting.
\newblock Facebook AI Research  (2020).
\newblock
  \urlprefix\url{https://ai.facebook.com/research/publications/neural-relational-autoregression-for-high-resolution-covid-19-forecasting}

\bibitem{Leclerc2009}
Leclerc, P.M., Matthews, A.P., Garenne, M.L.: Fitting the hiv epidemic in
  {Z}ambia: a two-sex micro-simulation model.
\newblock PloS one \textbf{4}(5), e5439 (2009)

\bibitem{Li2020}
Li, J., Vidyattama, Y., La, H.A., Miranti, R., Sologon, D.M.: The impact of
  {COVID-19} and policy responses on australian income distribution and
  poverty.
\newblock arXiv preprint \textbf{2009.04037} (2020).
\newblock \urlprefix\url{http://arxiv.org/abs/2009.04037}

\bibitem{Lipsitch2011}
Lipsitch, M., Finelli, L., Heffernan, R.T., Leung, G.M., Redd; for~the 2009
  H1N1 Surveillance~Group, S.C.: Improving the evidence base for decision
  making during a pandemic: the example of 2009 influenza a/h1n1.
\newblock Biosecurity and bioterrorism: biodefense strategy, practice, and
  science \textbf{9}(2), 89--115 (2011)

\bibitem{Liu2021}
Liu, S., Ni’mah, I., Menkovski, V., Mocanu, D.C., Pechenizkiy, M.: Efficient
  and effective training of sparse recurrent neural networks.
\newblock Neural Computing and Applications pp. 1--12 (2021)

\bibitem{Lloyd2004}
Lloyd, A.L., Jansen, V.A.: Spatiotemporal dynamics of epidemics: synchrony in
  metapopulation models.
\newblock Mathematical biosciences \textbf{188}(1-2), 1--16 (2004)

\bibitem{Lucas2019}
Lucas, B., Shifaz, A., Pelletier, C., O'Neill, L., Zaidi, N., Goethals, B.,
  Petitjean, F., Webb, G.I.: Proximity forest: an effective and scalable
  distance-based classifier for time series.
\newblock Data Mining and Knowledge Discovery \textbf{33}(3), 607--635 (2019).
\newblock \doi{10.1007/s10618-019-00617-3}

\bibitem{Malki2020}
Malki, Z., Atlam, E.S., Hassanien, A.E., Dagnew, G., Elhosseini, M.A., Gad, I.:
  Association between weather data and {COVID-19} pandemic predicting mortality
  rate: Machine learning approaches.
\newblock Chaos, Solitons \& Fractals \textbf{138}, 110137 (2020)

\bibitem{Melnick2020}
Melnick, E.R., Ioannidis, J.P.: Should governments continue lockdown to slow
  the spread of {COVID-19}?
\newblock BMJ \textbf{369} (2020)

\bibitem{Moore1994}
Moore, A.W., Lee, M.S.: Efficient algorithms for minimizing cross validation
  error.
\newblock In: Machine Learning Proceedings 1994, pp. 190--198. Elsevier (1994)

\bibitem{Mukherjee2021}
Mukherjee, H., Ghosh, S., Dhar, A., Obaidullah, S.M., Santosh, K., Roy, K.:
  Shallow convolutional neural network for covid-19 outbreak screening using
  chest x-rays.
\newblock Cognitive Computation pp. 1--14 (2021)

\bibitem{Mwalili2020}
Mwalili, S., Kimathi, M., Ojiambo, V., Gathungu, D., Mbogo, R.: {SEIR} model
  for {COVID-19} dynamics incorporating the environment and social distancing.
\newblock BMC Research Notes \textbf{13}(1), 1--5 (2020)

\bibitem{Opitz1996}
Opitz, D.W., Shavlik, J.W.: Actively searching for an effective neural network
  ensemble.
\newblock Connection Science \textbf{8}(3-4), 337--354 (1996)

\bibitem{Pal2020}
Pal, R., Sekh, A.A., Kar, S., Prasad, D.K.: Neural network based country wise
  risk prediction of {COVID-19}.
\newblock Applied Sciences \textbf{10}(18), 6448 (2020)

\bibitem{Peiris2004}
Peiris, J.S., Guan, Y., Yuen, K.Y.: Severe acute respiratory syndrome.
\newblock Nature medicine \textbf{10}(12), S88--S97 (2004)

\bibitem{Pettengill2020}
Pettengill, M.A., McAdam, A.J., Miller, M.B.: Can we test our way out of the
  {COVID-19} pandemic?
\newblock Journal of Clinical Microbiology \textbf{58}(11), e02225--20 (2020).
\newblock \doi{10.1128/JCM.02225-20}

\bibitem{Pichler2020}
Pichler, A., Pangallo, M., del Rio-Chanona, R.M., Lafond, F., Farmer, J.D.:
  Production networks and epidemic spreading: {H}ow to restart the {UK}
  economy?
\newblock arXiv preprint \textbf{2005.10585} (2020).
\newblock \urlprefix\url{http://arxiv.org/abs/2005.10585}

\bibitem{Quan2021}
Quan, D., Wong, L.L., Shallal, A., Madan, R., Hamdan, A., Ahdi, H., Daneshvar,
  A., Mahajan, M., Nasereldin, M., Van~Harn, M., et~al.: Impact of race and
  socioeconomic status on outcomes in patients hospitalized with {COVID-19}.
\newblock Journal of general internal medicine \textbf{36}(5), 1302--1309
  (2021)

\bibitem{Ray2020}
Ray, E.L., Wattanachit, N., Niemi, J., Kanji, A.H., House, K., Cramer, E.Y.,
  Bracher, J., Zheng, A., Yamana, T.K., Xiong, X., Woody, S., Wang, Y., Wang,
  L., Walraven, R.L., Tomar, V., Sherratt, K., Sheldon, D., Reiner, R.C.,
  Prakash, B.A., Osthus, D., Li, M.L., Lee, E.C., Koyluoglu, U., Keskinocak,
  P., Gu, Y., Gu, Q., George, G.E., Espa{\~n}a, G., Corsetti, S., Chhatwal, J.,
  Cavany, S., Biegel, H., Ben-Nun, M., Walker, J., Slayton, R., Lopez, V.,
  Biggerstaff, M., Johansson, M.A., Reich, N.G., on~behalf of~the {COVID-19}
  Forecast Hub~Consortium: Ensemble forecasts of coronavirus disease 2019
  ({COVID}-19) in the {U.S.}
\newblock medRxiv  (2020).
\newblock \doi{10.1101/2020.08.19.20177493}

\bibitem{Reich2019}
Reich, N.G., McGowan, C.J., Yamana, T.K., Tushar, A., Ray, E.L., Osthus, D.,
  Kandula, S., Brooks, L.C., Crawford-Crudell, W., Gibson, G.C., et~al.:
  Accuracy of real-time multi-model ensemble forecasts for seasonal influenza
  in the {US}.
\newblock PLoS computational biology \textbf{15}(11), e1007486 (2019)

\bibitem{Reich2020}
{Reich Lab - University of Massachusetts Amherst}: Data anomalies (2020).
\newblock
  \urlprefix\url{https://github.com/reichlab/covid19-forecast-hub/tree/master/data-anomalies}

\bibitem{Reinhart2021}
Reinhart, A., Brooks, L., Jahja, M., Rumack, A., Tang, J., Saeed, W.A., Arnold,
  T., Basu, A., Bien, J., Cabrera, {\'A}.A., Chin, A., Chua, E.J., Clark, B.,
  DeFries, N., Forlizzi, J., Gratzl, S., Green, A., Haff, G., Han, R., Hu,
  A.J., Hyun, S., Joshi, A., Kim, J., Kuznetsov, A., Motte-Kerr, W.L., Lee,
  Y.J., Lee, K., Lipton, Z.C., Liu, M.X., Mackey, L., Mazaitis, K., McDonald,
  D.J., Narasimhan, B., Oliveira, N.L., Patil, P., Perer, A., Politsch, C.A.,
  Rajanala, S., Rucker, D., Shah, N.H., Shankar, V., Sharpnack, J., Shemetov,
  D., Simon, N., Srivastava, V., Tan, S., Tibshirani, R., Tuzhilina, E.,
  Van~Nortwick, A.K., Ventura, V., Wasserman, L., Weiss, J.C., Williams, K.,
  Rosenfeld, R., Tibshirani, R.J.: An open repository of real-time covid-19
  indicators.
\newblock medRxiv  (2021).
\newblock
  \urlprefix\url{https://www.medrxiv.org/content/early/2021/07/16/2021.07.12.21259660}

\bibitem{delrio2020}
del Rio-Chanona, R.M., Mealy, P., Pichler, A., Lafond, F., Farmer, J.D.:
  {Supply and demand shocks in the {COVID-19} pandemic: an industry and
  occupation perspective}.
\newblock Oxford Review of Economic Policy \textbf{36}(Supplement 1), S94--S137
  (2020).
\newblock \doi{10.1093/oxrep/graa033}

\bibitem{Robishaw2021}
Robishaw, J.D., Alter, S.M., Solano, J.J., Shih, R.D., DeMets, D.L., Maki,
  D.G., Hennekens, C.H.: Genomic surveillance to combat covid-19: challenges
  and opportunities.
\newblock The Lancet Microbe  (2021)

\bibitem{Rodriguez2020}
Rodr{\'\i}guez, A., Tabassum, A., Cui, J., Xie, J., Ho, J., Agarwal, P.,
  Adhikari, B., Prakash, B.A.: Deepcovid: An operational deep learning-driven
  framework for explainable real-time {COVID-19} forecasting.
\newblock medRxiv  (2020).
\newblock \doi{10.1101/2020.09.28.20203109}

\bibitem{Radulescu2020}
Rǎdulescu, A., Williams, C., Cavanagh, K.: Management strategies in a
  {SEIR}-type model of {COVID-19} community spread.
\newblock Scientific reports \textbf{10}(1), 1--16 (2020)

\bibitem{Sak2014}
Sak, H., Senior, A., Beaufays, F.: Long short-term memory based recurrent
  neural network architectures for large vocabulary speech recognition.
\newblock arXiv preprint \textbf{1402.1128} (2014).
\newblock \urlprefix\url{http://arxiv.org/abs/1402.1128}

\bibitem{Sanche2020}
Sanche, S., Lin, Y.T., Xu, C., Romero-Severson, E., Hengartner, N., Ke, R.:
  High contagiousness and rapid spread of severe acute respiratory syndrome
  coronavirus 2.
\newblock Emerging infectious diseases \textbf{26}(7), 1470 (2020)

\bibitem{Schaffer1993}
Schaffer, C.: Selecting a classification method by cross-validation.
\newblock Machine Learning \textbf{13}(1), 135--143 (1993)

\bibitem{Shastri2021}
Shastri, S., Singh, K., Kumar, S., Kour, P., Mansotra, V.: Time series
  forecasting of {COVID-19} using deep learning models: {India-USA} comparative
  case study.
\newblock Chaos, Solitons \& Fractals \textbf{140}, 110227 (2020).
\newblock \doi{https://doi.org/10.1016/j.chaos.2020.110227}

\bibitem{Sherstinsky2020}
Sherstinsky, A.: Fundamentals of recurrent neural network ({RNN}) and long
  short-term memory ({LSTM}) network.
\newblock Physica D: Nonlinear Phenomena \textbf{404}, 132306 (2020)

\bibitem{deSouza2020}
de~Souza, W.M., Buss, L.F., da~Silva~Candido, D., Carrera, J.P., Li, S.,
  Zarebski, A.E., Pereira, R.H.M., Prete, C.A., de~Souza-Santos, A.A., Parag,
  K.V., et~al.: Epidemiological and clinical characteristics of the {COVID-19}
  epidemic in brazil.
\newblock Nature human behaviour \textbf{4}(8), 856--865 (2020)

\bibitem{Chen2017}
Sun, C., Shrivastava, A., Singh, S., Gupta, A.: Revisiting unreasonable
  effectiveness of data in deep learning era.
\newblock In: 2017 IEEE International Conference on Computer Vision (ICCV),
  vol. 2017-, pp. 843--852. IEEE (2017)

\bibitem{Surkova2020}
Surkova, E., Nikolayevskyy, V., Drobniewski, F.: False-positive {COVID-19}
  results: hidden problems and costs.
\newblock The Lancet Respiratory Medicine \textbf{8}(12), 1167--1168 (2020)

\bibitem{Tosepu2020}
Tosepu, R., Gunawan, J., Effendy, D.S., Lestari, H., Bahar, H., Asfian, P.,
  et~al.: Correlation between weather and {COVID-19} pandemic in jakarta,
  indonesia.
\newblock Science of the total environment \textbf{725}, 138436 (2020)

\bibitem{Vahedi2021}
Vahedi, B., Karimzadeh, M., Zoraghein, H.: Spatiotemporal prediction of
{COVID-19} cases using inter- and intra-county proxies of human interactions.
\newblock Nature Communications \textbf{12}, 6440 (2021).

\bibitem{Viboud2018}
Viboud, C., Sun, K., Gaffey, R., Ajelli, M., Fumanelli, L., Merler, S., Zhang,
  Q., Chowell, G., Simonsen, L., Vespignani, A., et~al.: The {RAPIDD} {EBOLA}
  forecasting challenge: Synthesis and lessons learnt.
\newblock Epidemics \textbf{22}, 13--21 (2018)

\bibitem{Walker2020}
Walker, P.G.T., Whittaker, C., Watson, O.J., Baguelin, M., Winskill, P.,
  Hamlet, A., Djafaara, B.A., Cucunubá, Z., Mesa, D.O., Green, W., Thompson,
  H., Nayagam, S., Ainslie, K.E.C., Bhatia, S., Bhatt, S., Boonyasiri, A.,
  Boyd, O., Brazeau, N.F., Cattarino, L., Cuomo-Dannenburg, G., Dighe, A.,
  Donnelly, C.A., Dorigatti, I., van Elsland, S.L., FitzJohn, R., Fu, H.,
  Gaythorpe, K.A.M., Geidelberg, L., Grassly, N., Haw, D., Hayes, S., Hinsley,
  W., Imai, N., Jorgensen, D., Knock, E., Laydon, D., Mishra, S.,
  Nedjati-Gilani, G., Okell, L.C., Unwin, H.J., Verity, R., Vollmer, M.,
  Walters, C.E., Wang, H., Wang, Y., Xi, X., Lalloo, D.G., Ferguson, N.M.,
  Ghani, A.C.: The impact of {COVID-19} and strategies for mitigation and
  suppression in low- and middle-income countries.
\newblock Science \textbf{369}(6502), 413--422 (2020).
\newblock \doi{10.1126/science.abc0035}

\bibitem{Wallinga2010}
Wallinga, J., van Boven, M., Lipsitch, M.: Optimizing infectious disease
  interventions during an emerging epidemic.
\newblock Proceedings of the National Academy of Sciences \textbf{107}(2),
  923--928 (2010)

\bibitem{Wang2020}
Wang, C., Liu, Z., Chen, Z., Huang, X., Xu, M., He, T., Zhang, Z.: The
  establishment of reference sequence for {SARS-CoV-2} and variation analysis.
\newblock Journal of medical virology \textbf{92}(6), 667--674 (2020)

\bibitem{Wang2020inference}
Wang, L., Didelot, X., Yang, J., Wong, G., Shi, Y., Liu, W., Gao, G.F., Bi, Y.:
  Inference of person-to-person transmission of covid-19 reveals hidden
  super-spreading events during the early outbreak phase.
\newblock Nature communications \textbf{11}(1), 1--6 (2020)

\bibitem{Watson2020}
Watson, J., Whiting, P.F., Brush, J.E.: Interpreting a {COVID-19} test result.
\newblock The British Medicial Journal \textbf{369} (2020).
\newblock \doi{10.1136/bmj.m1808}

\bibitem{Watts2005}
Watts, D.J., Muhamad, R., Medina, D.C., Dodds, P.S.: Multiscale, resurgent
  epidemics in a hierarchical metapopulation model.
\newblock Proceedings of the National Academy of Sciences \textbf{102}(32),
  11157--11162 (2005)

\bibitem{Wei2006}
Wei, Y., Pere, A., Koenker, R., He, X.: Quantile regression methods for
  reference growth charts.
\newblock Statistics in Medicine \textbf{25}(8), 1369--1382 (2006).
\newblock \doi{https://doi.org/10.1002/sim.2271}

\bibitem{Xu2020}
Xu, C., Yu, Y., Chen, Y., Lu, Z.: Forecast analysis of the epidemics trend of
  {COVID-19} in the {USA} by a generalized fractional-order {SEIR} model.
\newblock Nonlinear dynamics \textbf{101}(3), 1621--1634 (2020)

\bibitem{Yang2013}
Yang, J., Zeng, X., Zhong, S., Wu, S.: Effective neural network ensemble
  approach for improving generalization performance.
\newblock IEEE transactions on neural networks and learning systems
  \textbf{24}(6), 878--887 (2013)

\bibitem{Zeroual2020}
Zeroual, A., Harrou, F., Dairi, A., Sun, Y.: Deep learning methods for
  forecasting {COVID-19} time-series data: A comparative study.
\newblock Chaos, Solitons \& Fractals \textbf{140}, 110121 (2020)

\bibitem{Zhao2020}
Zhao, H., Lu, X., Deng, Y., Tang, Y., Lu, J.: {COVID-19}: asymptomatic carrier
  transmission is an underestimated problem.
\newblock Epidemiology \& Infection \textbf{148} (2020)

\bibitem{Zhou2020}
Zhou, T., Ji, Y.: Semiparametric bayesian inference for the transmission
  dynamics of covid-19 with a state-space model.
\newblock Contemporary Clinical Trials \textbf{97}, 106146 (2020)

\bibitem{Zou2020}
Zou, D., Wang, L., Xu, P., Chen, J., Zhang, W., Gu, Q.: Epidemic model guided
  machine learning for {COVID-19} forecasts in the {United States}.
\newblock medRxiv  (2020).

\end{thebibliography}
\end{document}